\documentclass[11pt,twoside]{article}
\input epsf 

\usepackage {jair,theapa}
\usepackage {url}
\usepackage {epsfig}
\usepackage {amsthm,amssymb,amsmath,amsfonts}

\jairheading{26}{2006}{247--287}{01/06}{07/06}
\ShortHeadings{The Landscape of Random Job Shop Scheduling Instances}
{Streeter \& Smith}
\firstpageno{247}

\newtheorem {theorem} {Theorem}
\newtheorem {lemma} {Lemma}
\newtheorem {corollary} {Corollary}
\newtheorem {proposition} {Proposition}
\newtheorem* {defjspinstance} {Definition (JSP instance)}
\newtheorem* {defjspschedule} {Definition (JSP schedule)}
\newtheorem* {defdisjgraph} {Definition (disjunctive graph)}
\newtheorem* {defdisjedge} {Definition (disjunctive edge)}
\newtheorem* {defdgdist} {Definition (disjunctive graph distance)}
\newtheorem* {defrandinstance} {Definition (random JSP instance)}
\newtheorem* {defpriorityrule} {Definition (priority rule)}
\newtheorem* {defrandsched} {Definition (random schedule)}
\newtheorem* {defrhobackbone} {Definition ($\rho\_backbone$)}
\newtheorem* {api} {Azuma's Perimetric Inequality (A.P.I.)}
\newtheorem* {whp} {Definition (whp)}
\newtheorem* {defprio0} {Definition (priority rule $\pi_0$)}
\newtheorem* {defprioinf} {Definition (priority rule $\pi_\infty$)}
\newtheorem* {defrdelta} {Definition ($(r, \delta)$-valley)}
\newtheorem* {defrdeltaplandscape} {Definition ($(r, \delta, p)$ landscape)}
\newtheorem* {defnr} {Definition (Neighborhood $\mathcal{N}_r$)}
\newtheorem* {deflopt} {Definition (local optimum $\mathcal {L} (S, \mathcal {N})$)}
\newtheorem* {defexactness} {Definition (neighborhood exactness)}
\newtheorem* {lemma2} {Lemma 2}
\newtheorem* {lemma3} {Lemma 3}
\newtheorem* {lemma4} {Lemma 4}

\newtheorem* {theorem6} {Theorem 6}

\def\argmax{\mathop{\rm arg\,max}}
\def\argmin{\mathop{\rm arg\,min}}

\numberwithin{equation}{section}

\begin{document}

\title{How the Landscape of Random Job Shop Scheduling Instances Depends on the Ratio of Jobs to Machines}

\author{\name Matthew J. Streeter \email matts@cs.cmu.edu \\
       \name Stephen F. Smith \email sfs@cs.cmu.edu \\
       \addr Carnegie Mellon University \\
       5000 Forbes Avenue, Pittsburgh, PA, 15213 USA }

\maketitle

\begin {abstract}
We characterize the search landscape of random instances of the job shop scheduling problem (JSP).  Specifically, we investigate how the expected values of (1) backbone size, (2) distance between near-optimal schedules, and (3) makespan of random schedules vary as a function of the job to machine ratio ($\frac {N} {M}$).  For the limiting cases $\frac {N} {M} \rightarrow 0$ and $\frac {N} {M} \rightarrow \infty$ we provide analytical results, while for intermediate values of $\frac {N} {M}$ we perform experiments.  We prove that as $\frac {N} {M} \rightarrow 0$, backbone size approaches 100\%, while as $\frac {N} {M} \rightarrow \infty$ the backbone vanishes.  In the process we show that as $\frac {N} {M} \rightarrow 0$ (resp. $\frac {N} {M} \rightarrow \infty$), simple priority rules almost surely generate an optimal schedule, providing theoretical evidence of an ``easy-hard-easy" pattern of typical-case instance difficulty in job shop scheduling.  We also draw connections between our theoretical results and the ``big valley" picture of JSP landscapes.
\end {abstract}

\section{Introduction}

\subsection{Motivations}

The goal of this work is to provide a picture of the typical landscape of a random instance of the job shop scheduling problem (JSP), and to determine how this picture changes as a function of the job to machine ratio ($\frac {N} {M}$).  Such a picture is potentially useful in (1) understanding how typical-case instance difficulty varies as a function of $\frac {N} {M}$ and (2) designing or selecting search heuristics that take advantage of regularities in typical instances of the JSP.

\subsubsection {Understanding instance difficulty as a function of $\frac {N} {M}$}

The job shop scheduling literature contains much empirical evidence that square JSPs (those with $\frac {N} {M} = 1$) are more difficult to solve than rectangular instances \cite {fisher63}.  This work makes both theoretical and empirical contributions toward understanding this phenomenon.  Empirically, we show that both random schedules and random local optima are furthest from optimality when $\frac {N} {M} \approx 1$.  Analytically, we prove that in the two limiting cases ($\frac {N} {M} \rightarrow 0$ and $\frac {N} {M} \rightarrow \infty$) there exist simple priority rules that almost surely produce an optimal schedule, providing theoretical evidence of an ``easy-hard-easy" pattern of instance difficulty in the JSP.  

\subsubsection {Informing the design of search heuristics}

Heuristics based on local search, for example tabu search \cite {glover97,nowicki96} and iterated local search (Louren\c{c}o, Martin, and St{\"u}tzle, 2003), have shown excellent performance on benchmark instances of the job shop scheduling problem \cite {jain98,jones98}.  In order to design an effective heuristic, one must (explicitly or implicitly) make assumptions about the search landscape of instances to which the heuristic will be applied.  For example, Nowicki and Smutnicki motivate the use of {\it path relinking} in their state-of-the-art {\it i}-TSAB algorithm by citing evidence that the JSP has a ``big valley" distribution of local optima \cite {nowicki05}.  One of the conclusions of our work is that the typical landscape of random instances can only be thought of as a big valley for values of $\frac {N} {M}$ close to 1; for larger values of $\frac {N} {M}$ (including values common in benchmark instances), the landscape breaks into many big valleys, suggesting that modifications to {\it i}-TSAB may allow it to better handle this case (we discuss $i$-TSAB further in \S 9.3). \nocite {lourenco03}

\subsection{Contributions}

The contributions of this paper are twofold.  First, we design a novel set of experiments and run these experiments on random instances of the JSP.  Second, we derive analytical results that confirm and provide insight into the trends suggested by our experiments.

The main contributions of our empirical work are as follows.

\begin {itemize}
	\item For low values of $\frac {N} {M}$, we show that low-makespan schedules are clustered in a small region of the search space and many attributes (i.e., directed disjunctive graph edges) are common to all low-makespan schedules. As $\frac {N} {M}$ increases, low-makespan schedules become dispersed throughout the search space and there are no attributes common to all low-makespan schedules.
	
	\item We introduce a statistic (neighborhood exactness) that can be used to quantitatively measure the ``smoothness" of a search landscape, and estimate the expected value of this statistic for random instances of the JSP.  These results, in combination with the results on clustering, suggest that the landscape of typical instances of the JSP can be described as a big valley only for low values of $\frac {N} {M}$; for high values of $\frac {N} {M}$ there are many separate big valleys.
\end {itemize}

For the limiting cases $\frac {N} {M} \rightarrow$ $0$ and $\frac {N} {M} \rightarrow$ $\infty$, we derive analytical results.  Specifically, we prove that

\begin {itemize}
	\item as $\frac {N} {M} \rightarrow$ $0$, the expected size of the backbone (i.e., the set of problem variables that have a common value in all global optima) approaches 100\%, while as $\frac {N} {M} \rightarrow$ $\infty$, the expected backbone size approaches 0\%; and
	
	\item as $\frac {N} {M} \rightarrow$ $0$ (resp. $\frac {N} {M} \rightarrow$ $\infty$), a randomly generated schedule will almost surely (a) be located ``close" in the search space to an optimal schedule and (b) have near-optimal makespan. 
\end {itemize}

\section{Related Work}

There are at least three threads of research that have conducted search space analyses related to the ones we conduct here.  These include literature on the ``big valley" distribution common to a number of combinatorial optimization problems, studies of backbone size in Boolean satisfiability, and a statistical mechanical analysis of the TSP.  We briefly review these three areas below, as well as relevant work on phase transitions and the ``easy-hard-easy" pattern of instance difficulty.

\subsection{The Big Valley}

The term ``big valley" originated in a paper by Boese et al. \citeyear {boese94} that examined the distribution of local optima in the Traveling Salesman Problem (TSP).  Based on a sample of local optima obtained by next-descent starting from random TSP tours, Boese calculated two correlations:

\begin {enumerate}
	\item the correlation between the cost of a locally optimal tour and its average distance to other locally optimal tours, and

	\item the correlation between the cost of a locally optimal tour and the distance from that tour to the best tour in the sample.
\end {enumerate}

The distance between two TSP tours was defined as the total number of edges minus the number of edges that are common to the two tours.  Based on the fact that both of these correlations were surprisingly high, Boese conjectured that local optima in the TSP are arranged in a ``big valley".  Adapted from the work of Boese et al. \citeyear {boese94}, Figure 1 gives ``an intuitive picture of the big valley, in which the set of local minima appears convex with one central global minimum" \cite {boese94}.  We offer a more formal definition of a big valley landscape in \S 6.

Boese's analysis has been applied to other combinatorial problems \cite {kim04}, including the permutation flow shop scheduling problem \cite {watson02,reeves98} and the JSP \cite {nowicki01}.  Correlations observed for the JSP are generally weaker than those observed for the TSP.  

In a related study, Mattfeld (1996) examined cost-distance correlations in the famous JSP instance {\tt ft10} \cite {beasley90} and  found evidence of a ``Massif Central\ldots where many near optimal solutions reside laying closer together than other local optima."  \S 4 contains related results on the backbone size of {\tt ft10}. \nocite {mattfeld96}

\begin {figure} \label {figbigvalley}
	\begin {center}
		\includegraphics [width=11cm,clip,trim=0cm 4.5cm 0cm 6cm] {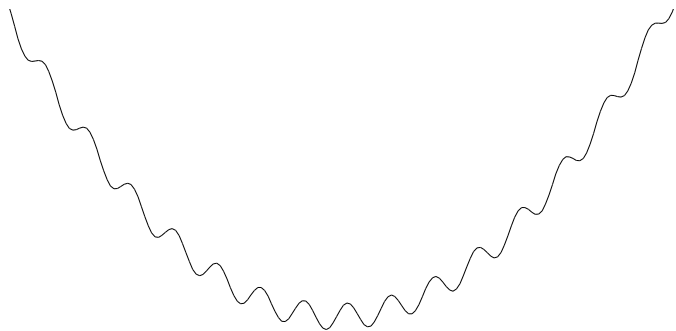}
	\end {center}
	\caption  {An intuitive picture of a ``big valley" landscape.} 
\end {figure}

\subsection{Backbone Size}

The {\it backbone} of a problem instance is the set of variables that are assigned a common value in all globally optimal solutions of that instance.  For example, in the Boolean satisfiability problem (SAT), the backbone is the set of variables that are assigned a fixed truth value in all satisfying assignments.  In the JSP, the backbone has been defined as the number of disjunctive edges (\S 3.2) that have a common orientation in all globally optimal schedules (a formal definition is given in \S 4).

There is a large literature on backbones in combinatorial optimization problems, including many empirical and analytical results \cite {slaney01,monasson99}.  In an analysis of problem difficulty in the JSP, Watson et al. \citeyear {watson01} present histograms of backbone size for random 6x6 (6 job, 6 machine) and 6x4 (6 job, 4 machine) JSP instances.  Summarizing experiments not reported in their paper, Watson et al. note that ``For [job:machine ratios] $>$ 1.5, the bias toward small backbones becomes more pronounced,  while for ratios $<$ 1, the bias toward larger backbones is further magnified.''  \S 4 generalizes these observations and  proves two theorems that give insight into why this phenomenon occurs.

\subsection{Statistical Mechanical Analyses}

A large and growing literature applies techniques from statistical mechanics to the analysis of combinatorial optimization problems \cite {martin01}.  At least one result obtained in this literature concerns clustering of low-cost solutions.  In a study of the TSP, M\'{e}zard and Parisi (1986) obtain an expression for the expected overlap (number of common edges) between random TSP tours drawn from a Boltzmann distribution.  They show that as the temperature parameter of the Boltzmann distribution is lowered (placing more probability mass on low-cost TSP tours), expected overlap approaches 100\%.  Though we do not use a Boltzmann weighting, \S 5 of this paper examines how expected overlap between random JSP schedules changes as more probability mass is placed on low-makespan schedules. \nocite {mezard86}

\subsection {Phase Transitions and the Easy-hard-easy Pattern}

Loosely speaking, a phase transition occurs in a system when the expected value of some statistic varies discontinuously (asymptotically) as a function of some parameter.  As an example, for any $\epsilon > 0$ it holds that  random instances of the 2-SAT problem are satisfiable with probability asymptotically approaching 1 when the clause to variable ratio ($\frac {m} {n}$) is $1-\epsilon$, but are satisfiable with probability approaching 0 when the clause to variable ratio is $1+\epsilon$.  A similar statement is conjectured to hold for 3-SAT; the critical value $k$ of $\frac {m} {n}$ (if it exists) must satisfy $3.42 \le k \le 4.51$ \cite {achlioptas04}.    

For some problems that exhibit phase transitions (notably 3-SAT), average-case instance difficulty (for typical solvers) appears to first increase and then decrease as one increases the relevant parameter, with the hardest instances appearing close to the threshold value \cite {cheeseman91,yokoo97}.  This phenomenon has been referred to as an ``easy-hard-easy" pattern of instance difficulty \cite {mammen97}.  In \S 7.4 we discuss evidence of an easy-hard-easy pattern of instance difficulty in the JSP, though (to our knowledge) it is not associated with any phase transition.
\\
\\
The results in \S \S 4-5 and the empirical results in \S 6 were previously presented in a conference paper \cite {streeter05}.

\section{The Job Shop Scheduling Problem}

\begin {figure} \label {figjsp}
	\begin {center}
		\includegraphics [width=11cm,clip,trim=0cm 5cm 0cm 1.5cm] {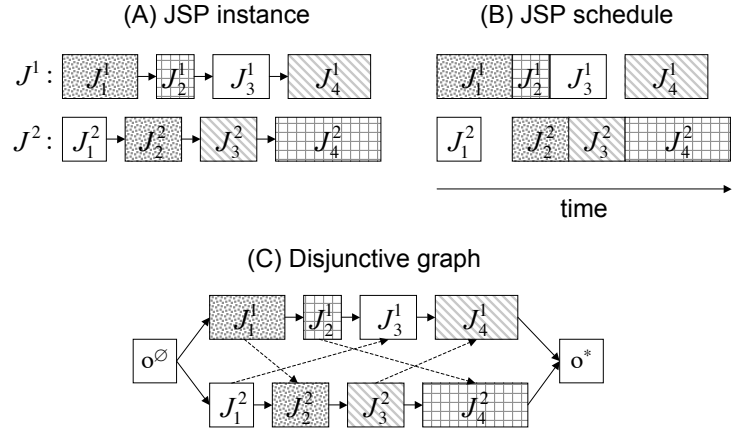}
	\end {center}
	\caption  {(A) A JSP instance, (B) a feasible schedule for the instance, and (C) the disjunctive graph representation of the schedule.  Boxes represent operations; operation durations are proportional to the width of a box; and the machine on which an operation is performed is represented by texture.  In (C), solid arrows represent conjunctive arcs and dashed arrows represent disjunctive arcs (arc weights are proportional to the duration of the operation the arc points out of).} 
\end {figure}

We adopt the notation $[n] \equiv \{1, 2, \ldots, n\}$.

\subsection {Problem Definition}

\begin {defjspinstance}
An $N$ by $M$ \emph {JSP instance} $I = \{$$J^1,$$J^2,$ \ldots, $J^N\}$ is a set of $N$ \emph {jobs} , where each job $J^k = (J^k_1, J^k_2, \ldots, J^k_M)$ is a sequence of $M$ \emph {operations}.  Each operation $o = J^k_i$ has an associated \emph {duration} $\tau(o) \in (0, \tau_{max}]$ and machine $m(o) \in [M]$.  We require that each job uses each machine exactly once (i.e., for each $J^k \in I$ and $\bar m \in [M]$, there is exactly one $i \in [M]$ such that $m(J^k_i) = \bar m$).  We define 
\begin {enumerate}
	\item $ops(I) \equiv \{J^k_i: k \in [N], i \in [M]\}$,
	\item $\tau(J^k) \equiv \sum_{i=1}^M \tau(J^k_i)$, and
	\item the \emph {job-predecessor} $\mathcal {J} (J^k_i)$ of an operation $J^k_i$ as
	\[ 
	\mathcal {J} (J^k_i) \equiv \left \{ 
		\begin {array} {l l}
			J^k_{i-1}	&	\mbox { if } i > 1 \\
			o^\emptyset		&	\mbox { otherwise }
		\end {array}
		\right . 
	\]
	where $o^\emptyset$ is a fictitious operation with $\tau(o^\emptyset) = 0$ and $m(o^\emptyset)$ undefined.
\end {enumerate}
\end {defjspinstance}

\begin {defjspschedule}
A JSP schedule for an instance $I$ is a function $S: ops(I) \rightarrow \Re_+$ that associates with each operation $o \in ops(I)$ a \emph {start time} $S(o)$ (operation $o$ is performed on machine $m(o)$ from time $S(o)$ to time $S(o)+\tau(o)$; preemption is not allowed).   We make the following definitions.

\begin {enumerate}
	\item The \emph {completion time} of an operation $o$ is $S^+(o)$ $\equiv$ $S(o) + \tau(o)$.
	\item The \emph {machine-predecessor} $\mathcal {M} (o)$ of an operation $o \in ops(I)$ is
	\[ 
	\mathcal {M} (o) \equiv \left \{ 
		\begin {array} {l l}
			\argmax_{\bar o \in O_{prev}(o)} S(\bar o) 	&	\mbox { if } O_{prev}(o) \neq \emptyset \\
			o^\emptyset		&	\mbox { otherwise. }
		\end {array}
		\right . 
	\]
	where $O_{prev}(o) = \{ \bar o \in ops(I): m(\bar o) = m(o), S(\bar o) < S(o) \}$ is the set of operations scheduled to run before $o$ on $o$'s machine.
	\item $S$ is a \emph {feasible} schedule if $S(o) \ge \max (S^+(\mathcal {J} (o)), S^+(\mathcal {M} (o)))$ $\forall o \in ops(I)$.
	\item The quantity
		\begin {displaymath}
			\ell(S) \equiv \max_{o \in ops(I)} S^+(o)
		\end {displaymath}
		is called the \emph {makespan} of $S$.  
\end {enumerate}

\end {defjspschedule}

We consider the makespan-minimization version of the JSP, in which the goal is to find a schedule that minimizes the makespan.  

For the remainder of the paper, whenever we refer to a JSP schedule $S$ we shall adopt the convention that $S(o^\emptyset) = 0$ and we shall assume that
\begin {equation} \label {semiactive}
S(o) = \max (S^+(\mathcal {J} (o)), S^+(\mathcal {M} (o))) \mbox {   } \forall o \in ops(I)
\end {equation}
(i.e., $S$ is a so-called \emph {semi-active} schedule, French, 1982).  In other words, we ignore schedules with superfluous idle time at the start of the schedule or between the end of one operation and the start of another. \nocite {french82}

Figure 2 (A) and (B) depict, respectively, a JSP instance and a feasible schedule for that instance.  

\subsection {Disjunctive Graphs}

A schedule satisfying \eqref {semiactive} can be uniquely represented by a weighted, directed graph called its \emph {disjunctive graph}.  In the disjunctive graph representation of a schedule $S$ for a JSP instance $I$, each operation $o \in ops(I)$ is a vertex and a directed edge $(o_1, o_2)$ indicates that operation $o_1$ completes before $o_2$ starts.

\begin {defdisjgraph}
The disjunctive graph $G = \mathcal{G}(I, S)$ of a schedule $S$ for a JSP instance $I$ is the weighted, directed graph  $G = (V,\vec E,w)$ defined as follows.  
	\begin {itemize}
		\item $V = ops(I) \cup \{o^\emptyset, o^*\}$, where $o^*$ (like $o^\emptyset$) is a fictitious operation with $\tau(o^*) = 0$ and $m(o^*)$ undefined.
		\item $\vec E = \vec C \cup \vec D$, where
		\begin {itemize}
			\item 
			$\vec C = \left \{ (\mathcal {J} (o), o): o \in ops(I) \right \} \cup \left \{ (J^k_M, o^*): k \in [N] \right \}$ is called the set of \emph {conjunctive arcs} (which specify that $o$ cannot start until $\mathcal {J}(o)$ completes), and
			\item $\vec D = \left \{ (o_1, o_2): \{o_1,o_2\} \subseteq ops(I) , m(o_1) = m(o_2), S(o_1) < S(o_2) \right \}$ is called the set of \emph {disjunctive arcs} (which specify, for each pair of operations performed on the same machine, which of the two operations is to be performed first).
		\end {itemize}
		\item $w((o_1, o_2)) = \tau(o_1)$.
	\end {itemize}
\end {defdisjgraph}

Figure 2 (C) depicts the disjunctive graph for the schedule depicted in Figure 2 (B).  The connection between a schedule and its disjunctive graph is established by the following proposition \cite {roy64}.

\begin {proposition}
	Let $S$ be a feasible schedule for $I$ satisfying \eqref {semiactive}, and let $G = \mathcal{G}(I, S)$ be the corresponding disjunctive graph.  Then $\ell(S)$ is equal to the length of the longest weighted path from $o^\emptyset$ to $o^*$ in $G$.
\end {proposition}
\begin {proof}
For any operation $o$, let $L(o)$ denote the length of the longest weighted path from $o^\emptyset$ to $o$ in $G$.  It suffices to show that for any $o \in ops(I)$, $S(o) = L(o)$.  This follows by induction on the number of edges in the path, with the base case $S(o^\emptyset) = L(o^\emptyset) = 0$.
\end {proof}

The undirected version of a disjunctive arc is called a \emph {disjunctive edge}.

\begin {defdisjedge}
	Let $I$ be a JSP instance.  A \emph {disjunctive edge} is a set $\{o_1, o_2\} \subset ops(I)$ with $m(o_1) = m(o_2)$.  We define the following notation.
\begin {itemize}
	\item $E(I)$ is the set of disjunctive edges for $I$.
	\item Let $S$ be a schedule for $I$ and let $e = \{o_1, o_2\}$ be a disjunctive edge.  We denote by $\vec e(S)$ the unique arc in $\{ (o_1,o_2), (o_2,o_1)  \}$ that appears in the disjunctive graph $\mathcal{G}(I,S)$ (this arc is called the \emph {orientation} of $e$ in $S$).
\end {itemize}
\end {defdisjedge}

We measure the distance between two schedules $S_1$ and $S_2$ for a JSP instance $I$ by counting the number of disjunctive edges that are oriented in opposite directions in $\mathcal{G}(I,S_1)$ and $\mathcal{G}(I,S_2)$.

\begin {defdgdist}
The \emph {disjunctive graph distance} $\| S_1 - S_2\|$ between two schedules $S_1$ and $S_2$ for a JSP instance $I$ is defined by
		\begin {displaymath}
			\| S_1 - S_2\| \equiv | \{ e \in E(I): \vec e(S_1) \neq \vec e(S_2) \} | \mbox { .}
		\end {displaymath}
\end {defdgdist}

\subsection {Random Schedules and Instances}

We define a uniform distribution over JSP instances as follows.  Our distribution is identical to the one used by Taillard (1993). \nocite {taillard93}

\begin {defrandinstance}
A random $N$ by $M$ JSP instance $I$ is generated as follows.
	\begin {enumerate}
		\item Let $\phi_1, \phi_2, \ldots, \phi_N$ be random permutations of $[M]$.
		\item Let $G$ be a probability distribution over $(0, \tau_{max}]$ with mean $\mu$ and variance $\sigma^2 > 0$.
		\item Define $I = \{J^1, J^2, \ldots, J^N \}$, where $m(J^k_i) = \phi_k(i)$ and each $\tau(J^k_i)$ is drawn (independently at random) from $G$.
	\end {enumerate}
\end {defrandinstance}

Note that this definition (and likewise, our theoretical results) assumes a maximum operation duration $\tau_{max}$, but makes no assumptions about the form of the distribution of operation durations.  For the empirical results reported in this paper, we choose operation durations from a uniform distribution over $\{ 1, 2, \ldots, 100 \}$.  

Our proofs will frequently make use of \emph {priority rules}.  A priority rule is a greedy schedule-building  algorithm that assigns a priority to each operation and, at each step of the greedy algorithm, assigns the earliest possible start time to the operation with minimum priority.

\begin {defpriorityrule}
A \emph {priority rule} $\pi$ is a function that, given an instance $I$ and an operation $o \in ops(I)$, returns a priority $\pi(I, o) \in \Re$.  The schedule $S = \mathcal {S} (\pi, I)$ associated with $\pi$ is defined by the following procedure.
\begin {enumerate}
	\item $Unscheduled \leftarrow ops(I)$, $S(o^\emptyset) \leftarrow 0$.
	\item While $|Unscheduled| > 0$ do:
	\begin {enumerate}
		\item $Ready \leftarrow \{ o \in Unscheduled: \mathcal {J} (o) \notin Unscheduled \}$.  
		\item $\bar o \leftarrow$ the element of $Ready$ with least priority.
		\item $S(\bar o) \leftarrow \max (S^+(\mathcal {J} (\bar o)), S^+(\mathcal {M} (\bar o)))$.
		\item Remove $\bar o$ from $Unscheduled$.
	\end {enumerate}
\end {enumerate}
	A priority rule is called \emph {instance-independent} if, for any $N$ by $M$ JSP instance $I$ and integers $k \in [N]$, $i \in [M]$, the value $\pi(I, J^k_i)$ depends only on $k$, $i$, $N$, and $M$.
\end {defpriorityrule}

We obtain a random schedule by assigning random priorities to each operation.  The resulting distribution is equivalent to the one used by Mattfeld (1996). \nocite {mattfeld96} 

\begin {defrandsched}
A \emph {random schedule} for an $N$ by $M$ JSP instance $I$ is generated by performing the following steps.
\begin {enumerate}
	\item Create a list $L$ containing $M$ occurrences of the integer $k$ for each $k \in [N]$ (we think of the $M$ occurrences of $k$ as representing the operations in the job $J^k$).  
	\item Shuffle $L$ (obtaining each permutation with equal probability). 
	\item Return the schedule $\mathcal {S} (\pi_{rand}, I)$ where $\pi_{rand} (I,J^k_i) =$ the index of the $i^{th}$ occurrence of $k$ in $L$.
\end {enumerate}
\end {defrandsched}

\section{Number of Common Attributes as a Function of Makespan}

The \emph {backbone} of a JSP instance is the set of disjunctive edges that have a common orientation in all schedules whose makespan is globally optimal.  For $\rho \ge 1$, we define the $\rho\_backbone$ to be the set of disjunctive edges that have a common orientation in all schedules whose makespan is within a factor $\rho$ of optimal (a related definition appears in Slaney \& Walsh, 2001). \nocite {slaney01}

\begin {defrhobackbone}
Let $I$ be a JSP instance with optimal makespan $\ell_{min}(I)$.  For $\rho \ge 1$, let $\rho\_opt(I) \equiv \{S: \ell(S) \le \rho\cdot \ell_{min}(I)\}$ be the set of schedules whose makespan is within a factor $\rho$ of optimal.  Then
	\begin {displaymath}
		\rho\_backbone(I) \equiv \{ e \in E(I): \vec e(S_1) = \vec e(S_2) \mbox{ } \forall \{S_1,S_2\} \subseteq \rho\_opt(I) \} \mbox { .}
	\end {displaymath}
\end {defrhobackbone}

In this section we compute the expected value of $| \rho\_backbone |$ as a function of $\rho$ for random {\it N} by {\it M} JSP instances, and examine how the shape of this curve changes as a function of $\frac {N} {M}$.  

\subsection{Computing the $ \rho\_backbone$ }

To compute the $\rho\_backbone$ we use the following proposition.
\begin {proposition}
Let $I$ be a JSP instance with optimal makespan $\ell_{min}(I)$.  Let $e = \{o_1, o_2\}$ be a disjunctive edge with orientations $a_1 = (o_1, o_2)$ and $a_2 = (o_2, o_1)$.  For any disjunctive arc $a$, let $\ell_{min}(I | a)$ denote the optimum makespan among schedules whose disjunctive graph contains the arc $a$.  Then
\begin {displaymath}
e \in \rho\_backbone(I) \Leftrightarrow \max \left \{ \ell_{min}(I | a_1), \ell_{min}(I | a_2) \right \} > \rho \cdot \l_{min}(I) \mbox { .}
\end {displaymath}
\end {proposition}
\begin {proof}
If $e \in \rho\_backbone$, then $e$ must have a common orientation (say $a_1$) in all schedules $S$ with $\ell(S) \le \rho \cdot \ell_{min}(I)$, which implies $\ell_{min}(I | a_2) > \rho \cdot \ell_{min}(I)$.  If $e \notin \rho\_backbone$, then there must be some $\{S_1, S_2\} \subseteq \rho\_opt(I)$ with $\vec e (S_1) = a_1$ and $\vec e(S_2) = a_2$, which implies $\max \{ \ell_{min}(I | a_1),\ell_{min}(I | a_2) \} \le \rho \cdot \ell_{min} (I).$ 
\end {proof}

Thus to compute $\rho\_backbone(I)$ we need only to compute $\ell_{min}(I | a)$ for the $2 M {N \choose 2}$ possible choices of $a$.  Given a disjunctive arc $a$, we compute $\ell_{min}(I | a)$ using branch and bound.  In branch and bound algorithms for the JSP, nodes in the search tree represent choices of orientations for a subset of the disjunctive edges.  By constructing a root search tree node that has $a$ as a fixed arc, we can determine $\ell_{min}(I | a)$.  We use a branch and bound algorithm due to Brucker et al. \citeyear {brucker94} because it is efficient and because the code for it is freely available via ORSEP \cite {brucker92}.

Computing $\ell_{min}(I | a)$ for the $2 M {N \choose 2}$ possible choices of $a$ requires only $1 + M {N \choose 2}$ runs of branch and bound.  The first run is used to find a globally optimal schedule, which gives the value of $\ell_{min} (I | a)$ for $M {N \choose 2}$ possible choices of $a$ (namely, the $M {N \choose 2}$ disjunctive arcs that are present in the globally optimal schedule).  A separate run is used for each of the $M {N \choose 2}$ remaining choices of $a$.

Figure 3 graphs the fraction of disjunctive edges that belong to the $\rho\_backbone$ as a function of $\rho$ for instance {\tt ft10} (a 10 job, 10 machine instance) from the OR library \cite {beasley90}.  Note that by definition the curve is non-increasing with respect to $\rho$, and that the curve is exact for all $\rho$.  It is noteworthy that among schedules whose makespan is within a factor 1.005 of optimal, 80\% of the disjunctive edges have a fixed orientation.  We will see that this behavior is typical of JSP instances with $\frac {N} {M} = 1$.

\begin {figure}
	\begin {center}
		\includegraphics [width=8.5cm,clip,trim=1cm 1cm 0cm 0cm] {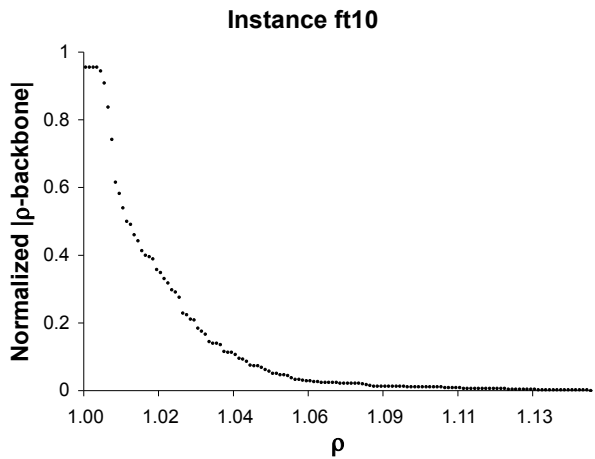}
	\end {center}
	\caption  {Normalized $| \rho\_backbone |$ as a function of $\rho$ for OR library instance {\tt ft10}.}
\end {figure}

\subsection{Results}

We plotted $| \rho\_backbone|$ as a function of $\rho$ for all instances in the OR library having 10 or fewer jobs and 10 or fewer machines.  The results are available online \cite {streeter05s}.  Inspection of the graphs revealed that the shape of the curve is largely a function of the job:machine ratio.  To investigate this further, we repeat these experiments on a large number of randomly generated JSP instances.

We use randomly generated instances with 7 different combinations of $N$ and {\it M} to study instances with $\frac {N} {M}$ equal to 1, 2, or 3.  For $\frac {N} {M} = 1$ we use 6x6, 7x7, and 8x8 instances; for $\frac {N} {M} = 2$ we use 8x4 and 10x5 instances; and for $\frac {N} {M} = 3$ we use 9x3 and 12x4 instances.  We generate 1000 random instances for each combination of $N$  and $M$.

Figure 4 parts (A), (B), and (C) graph the expected fraction of edges belonging to the $\rho$-backbone as a function of $\rho$ for each combination of $N$ and $M$, grouped according to $\frac {N} {M}$.  Figure 4 (D) compares the curves for different values of $\frac {N} {M}$, and plots the 0.25 and 0.75 quantiles.  For the purposes of this study the two most important observations about Figure 4 are as follows.
	
\begin {figure} [p]
	\begin {center}
		\includegraphics [width=7.5cm,clip,trim=1cm 0.5cm 1.5cm 0cm] {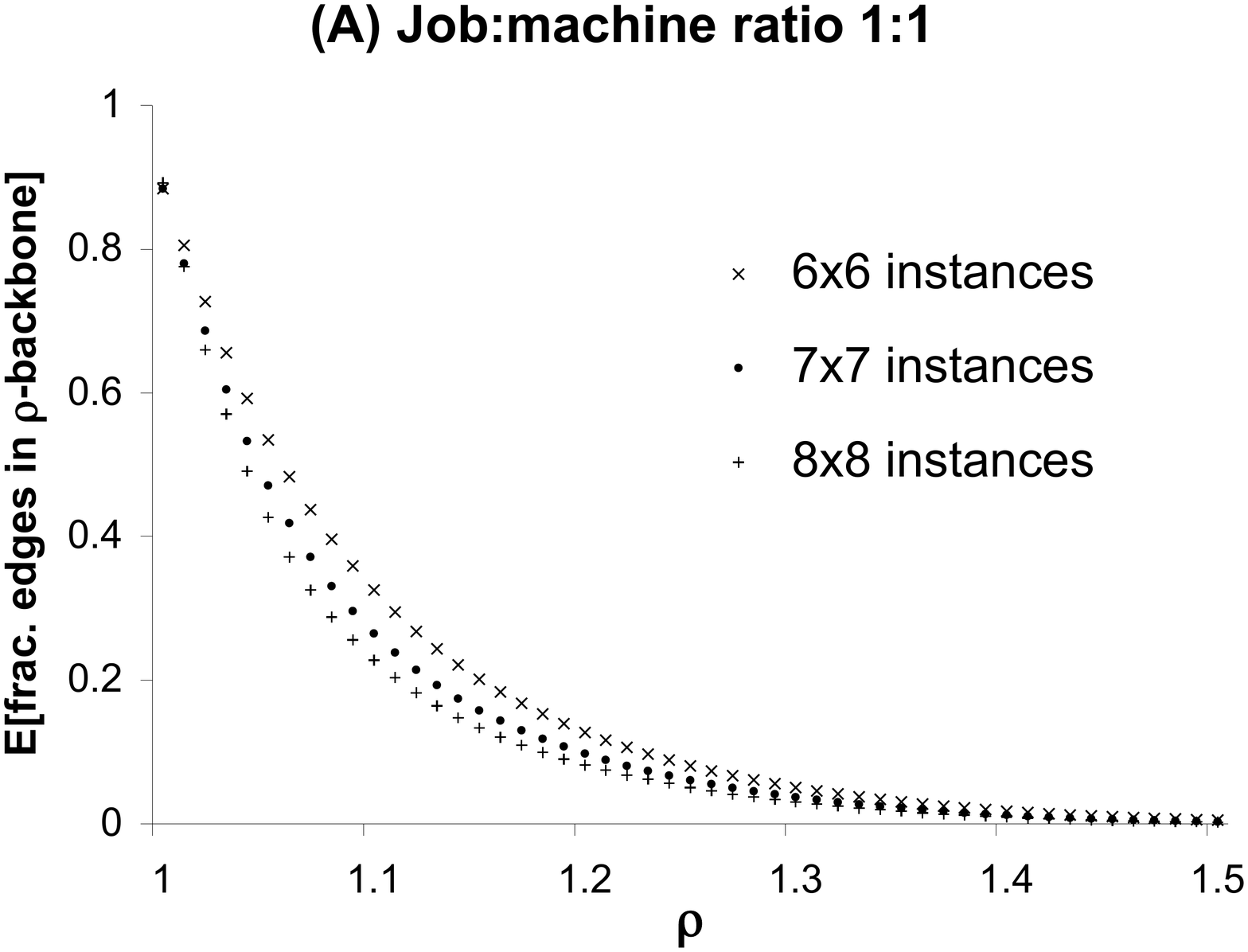} \includegraphics [width=7.5cm,clip,trim=1cm 0.5cm 1.5cm 0cm] {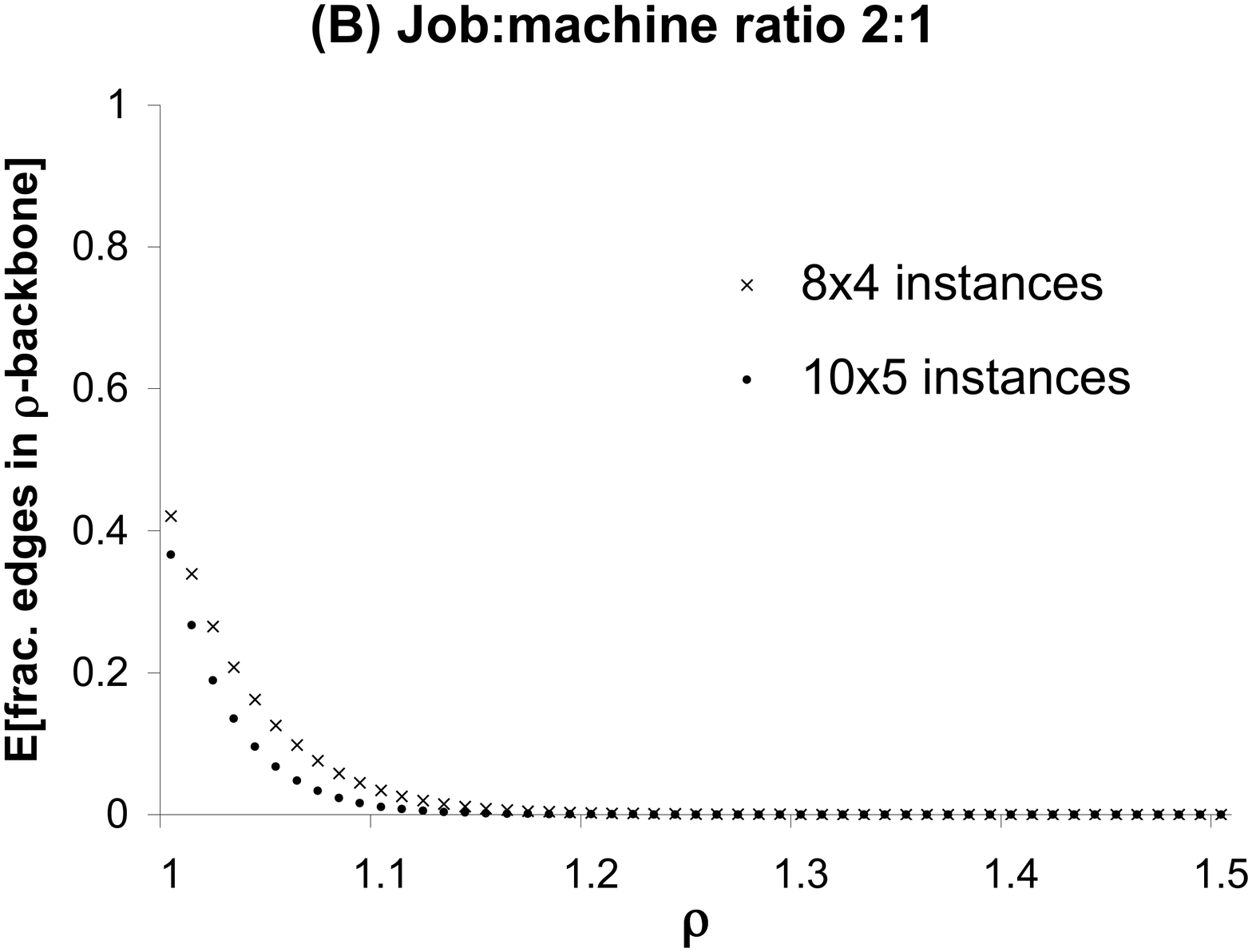} \\
		\includegraphics [width=7.5cm,clip,trim=1cm 0.5cm 1.5cm 0cm] {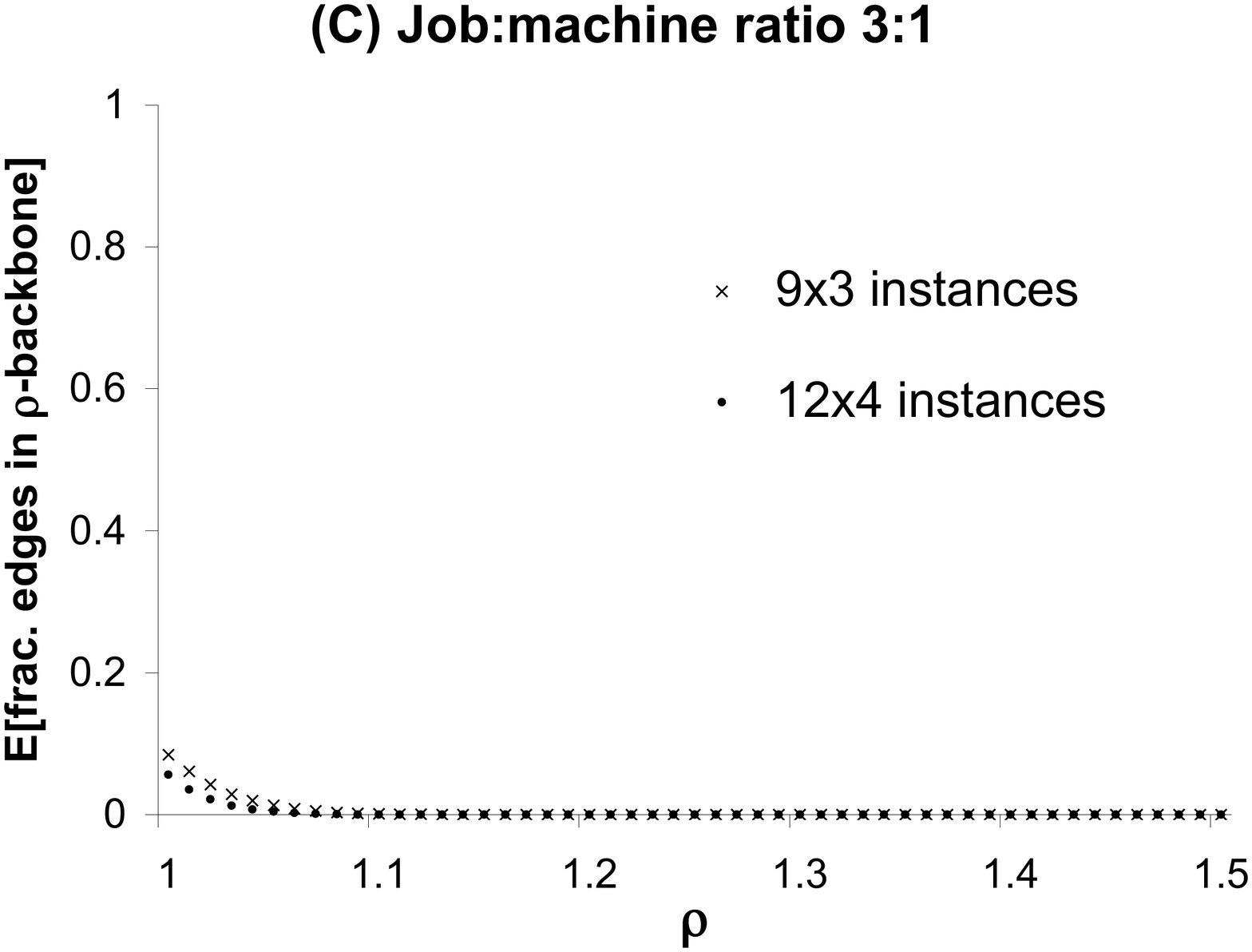} \includegraphics [width=7.5cm,clip,trim=1cm 0.5cm 1.5cm 0cm] {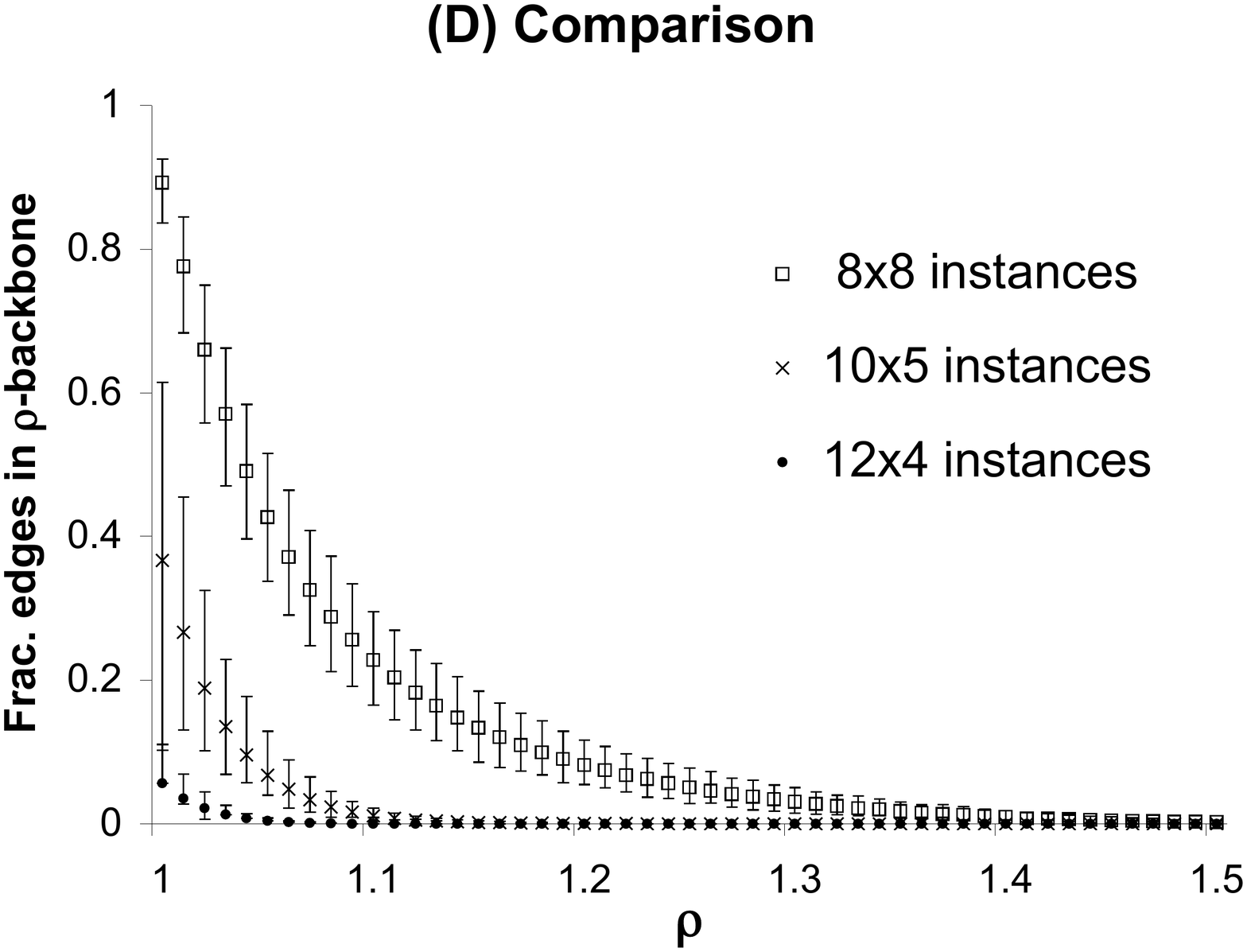}
	\end {center}
	\caption  {Expected fraction of edges in $\rho$-backbone as a function of $\rho$ for random JSP instances.  Graphs (A), (B), and (C) depict curves for random instances with $\frac {N} {M}$ = 1, 2, and 3, respectively.  Graph (D) compares the curves depicted in (A), (B), and (C) (only the curves for the largest instance sizes are shown in (D)).  In (D), top and bottom error bars represent 0.75 and 0.25 quantiles, respectively.}
\end {figure}

\begin {itemize}
	\item The curves depend on both the size of the instance (i.e., $NM$) and the shape (i.e., $\frac {N} {M}$).  Of these two factors, $\frac {N} {M}$ has by far the stronger influence on the shape of the curves.

	\item For {\it all} values of $\rho$, the expected fraction of edges belonging to the $\rho\_backbone$ decreases as $\frac {N} {M}$ increases.  
\end {itemize}

\subsection{Analysis}

We now give some insight into Figure 4 by analyzing two limiting cases.  We prove that as $\frac {N} {M}$$\rightarrow$$0$, the expected fraction of disjunctive edges that belong to the backbone approaches 1, while as $\frac {N} {M}$$\rightarrow$$\infty$ this expected fraction approaches 0.

Intuitively, what happens is as follows.  As $\frac {N} {M}$$\rightarrow$$0$ (i.e., $N$ is held constant and $M$$\rightarrow$$\infty$) each of the jobs becomes very long.  Individual disjunctive edges then represent precedence relations among operations that should be performed very far apart in time.  For example, if there are 10,000 machines (and so each job consists of 10,000 operations), a disjunctive edge might specify whether operation 1,200 of job $A$ is to be performed before operation 8,500 of job $B$.  Clearly, waiting for job $B$ to complete 8,500 of its operations before allowing job $A$ to complete 12\% of its operations is likely to produce an inefficient schedule.  Thus, orienting a single disjunctive edge in the ``wrong" direction is likely to prevent a schedule from being optimal, and so any particular edge will likely have a common orientation in all globally optimal schedules.

In contrast, when $\frac {N} {M}$$\rightarrow$$\infty$, it is the workloads of the machines that become very long.  The order in which the jobs are processed on a particular machine does not matter much as long as the machine with the longest workload is kept busy, and so the fact that a particular edge is oriented a particular way is unlikely to prevent a schedule from being optimal.  All of this is formalized below.  

We will make use of the following well-known definition.

\begin {whp}
A sequence of events $\xi_{n}$ occurs with high probability (whp) if $\lim_{n \to\infty}$ $ 
\mathbb{P} [\xi_{n}] = 1$.
\end {whp}

Lemma \ref {lem:backbone:0} and Theorem \ref {backbone:0} show that for constant $N$, a randomly chosen edge of a random $N$ by $M$ JSP instance will be in the backbone whp (as $M$$\rightarrow$$\infty$).  Lemma \ref {lem:backbone:inf} and Theorem \ref {backbone:inf} show that for constant $M$, a randomly chosen edge of a random $N$ by $M$ JSP instance will not be in  the backbone whp (as $N$$\rightarrow$$\infty$).

\begin {lemma} \label {lem:backbone:0}
Let $I$ be a random $N$ by $M$ JSP instance, and let $S = \mathcal {S} (\pi, I)$ be the schedule for $I$ obtained using some instance-independent priority rule $\pi$.  For an arbitrary job $J \in I$, define $\Delta^S_J \equiv S^+(J_M) -  \tau(J)$.  Then $\mathbb {E} [\Delta^S_J]$ is $O(N)$.
\end {lemma}

\begin {proof}

We assume $N=2$ and $M>1$.  The generalization to larger $N$ is straightforward, while the cases $N=1$ and $M=1$ are trivial.  Let $I = \{J^1, J^2\}$ and let $J = J^1$.  

Let $T = (\bar o_1, \bar o_2, \ldots, \bar o_{N M})$ be the sequence of operations selected from $Ready$ (in line 2(b) of the definition of a priority rule in \S 3.3) in constructing $S$. We say that an operation $J^1_i$ \emph {overlaps} with an operation $J^2_j$ if
\begin {enumerate}
	\item $J^2_j$ appears before $J^1_i$ in $T$, and
	\item $[S(J^2_j), S^+(J^2_j)] \cap [ S^+(J^1_{i-1}), S^+(J^1_{i-1}) + \tau(J^1_i) ] \neq \emptyset$ .
\end {enumerate}
If additionally $m(J^1_i) = m(J^2_j)$, we say that $J^1_i$ \emph {contends} with $J^2_j$.  Intuitively, if $o \equiv J^1_i$ overlaps with $o' \equiv J^2_j$ then the start time of $o$ \emph {might} have been delayed because $o$'s machine was being used by $o'$.  If $o$ contends with $o'$, then the start time of $o$ actually \emph {was} delayed.

Let $\theta_{i,j}$ (resp. $\delta_{i,j}$) be an indicator for the event that $J^1_i$ overlaps (resp. contends) with $J^2_j$.  Let $C_i \equiv \{J^2_j: \theta_{i,j} = 1 \}$ be the set of operations in $J^2$ that $J^1_i$ overlaps with.  Then $| C_i \cap \bigcup_{i' > i} C_{i'} | \le 1$.  Thus
\begin {equation} \label {eq:overlap}
	\sum_i { | C_i| } 
	= \sum_i { | C_i \setminus \bigcup_{i' > i} C_{i'} | } + \sum_i { | C_i \cap \bigcup_{i' > i} C_{i'} | } 
	\le 2 M \mbox { .}
\end {equation}

Let $\bar I = I_{N,M-1}$ be a random $N$ by $M-1$ JSP instance, and define $\bar \theta_{i,j}$, $\bar \delta_{i,j}$, and $\bar C_i$ analogously to the above.  Then for $i,j \le M-1$, 
\begin {displaymath}
\mathbb {P} \left [ \theta_{i,j} = 1 | m(J^1_i) = m(J^2_j) \right ] = \mathbb {P} \left [ \bar \theta_{i,j} = 1 \right ] \mbox { .}
\end {displaymath}
This is true because $\mathbb {P} [\theta_{i,j} = 1]$ is a function of the joint distribution of the operations in the set $\{ J^1_{i'} : i' < i \} \cup \{ J^2_{j'} : j' < j \}$; and, as far as this joint distribution is concerned, conditioning on the event $m(J^1_i) = m(J^2_j)$ is like deleting the operations that use the machine $m(J^1_i)$.  

Thus $\mathbb {E} \left [ \delta_{i,j} \right ] = \mathbb {P} \left [ \delta_{i,j} = 1 \right ] =  \frac {1} {M} \mathbb {P} \left [ \theta_{i,j} = 1 | m(J^1_i) = m(J^2_j) \right ]  = \frac {1} {M} \mathbb {P} \left [ \bar \theta_{i,j} = 1 \right ] =  \frac {1} {M} \mathbb {E} \left [ \bar \theta_{i,j} \right ]$.  Therefore, 
\[
\begin {array} {l l}
	\sum_{i=1}^M { \sum_{j=1}^M \mathbb {E} [ \delta_{i,j} ] }  & \le 2 + \sum_{i=1}^{M-1} { \sum_{j=1}^{M-1} \mathbb {E} [ \delta_{i,j} ] } \\
	& = 2 + \frac {1} {M} \sum_{i=1}^{M-1} { \sum_{j=1}^{M-1} \mathbb {E} [ \bar \theta_{i,j} ] } \\
	& = 2 + \frac {1} {M} \sum_{i=1}^{M-1} \mathbb {E} [ |\bar C_i|]  \\
	& \le 4
\end {array}
\]
where in the last step we have used \eqref {eq:overlap}.  It follows that $\mathbb {E} [\Delta^S_J] \le 4 \tau_{max}$ ($\tau_{max}$ is the maximum operation duration defined in \S 3).  When we consider arbitrary $N$, we get $\mathbb {E} [\Delta^S_J] \le 4 \tau_{max} (N-1)$. 

\end {proof}

As a corollary of Lemma \ref {lem:backbone:0}, we can show that a simple priority rule ($\pi_0$) almost surely generates an optimal schedule in the case $\frac {N} {M} \rightarrow 0$.

\begin {defprio0}
Given an $N$ by $M$ JSP instance $I$, let $k^* = \argmax_{k \in [N]}$ ${\tau(J^k)}$ be the index of the longest job.  The priority rule $\pi_0$ first schedules the operations in $J^{k^*}$, then schedules the remaining operations in a fixed order.
\begin {displaymath}
	\pi_0 (I,J^k_i) = \left \{ \begin {array} {l l} 
		i & \mbox { if } k=k^* \\
		M k+i & \mbox { otherwise.}
	\end {array} \right .
\end {displaymath}  
\end {defprio0}

\begin {corollary} \label {optmakespan:0}
Let $I$ be a random $N$ by $M$ JSP instance.  Then for fixed $N$, it holds whp (as $M \rightarrow \infty$) that the schedule $S = \mathcal {S} (\pi_0, i)$ is optimal and has makespan $\ell(S) = \max_{k \in [N]} \tau(J^k)$.
\end {corollary}

\begin {proof}
Define the priority rule $\pi_{\bar k}$ by $\pi_{\bar k}(I,J^k_i) = i$ if $k=\bar k$; $M k+i$ otherwise.  Then $\pi_{\bar k}$ is instance-independent, and $\pi_0$ is equivalent to $\pi_{k^*}$.  Thus for any $J \in I$ we have

\begin {displaymath}
	\mathbb {E} [\Delta^{\pi_{0}}_{J}] \le \sum_{k} \mathbb {E} [\Delta^{\pi_k}_{J}] = O(N^2)
\end {displaymath}

where we define $\Delta^{\pi}_J \equiv \Delta^{\mathcal {S} (\pi, I)}_J$, and the second step uses Lemma \ref {lem:backbone:0}.  By Markov's inequality, $\Delta^{\pi_0}_{J} < M^{\frac {1} {4}}$ $\forall J \in I$ whp.  By the Central Limit Theorem, each $\tau(J)$ is asymptotically normally distributed with mean $\mu M$ and standard deviation $\sigma \sqrt M$.  It follows that whp, $\tau(J^{k^*}) - \tau(J^k) > M^{\frac {1} {4}}$ $\forall k \neq k^*$.  This implies $\ell(S) = \tau(J^{k^*})$.  Because $\tau(J^{k^*})$ is a lower bound on the makespan of any schedule, the corollary follows.
\end {proof}


\begin {theorem} \label {backbone:0}
Let $I$ be a random $N$ by $M$ JSP instance, and let $e$ be a randomly selected element of $E(I)$.  Then for fixed $N$, it holds whp (as $M$$\rightarrow$$\infty$) that $e \in 1\_backbone(I)$.
\end {theorem}

\begin {proof}
Let $e = \{J_i, J'_j\}$ with $i \le j$ and let $a = (J'_j, J_i)$.  By Proposition 1 and Corollary \ref {optmakespan:0}, it suffices to show that whp, all disjunctive graphs containing $a$ contain a path from $o^\emptyset$ to $o^*$ with weighted length $> \max_{k \in [N]} \tau(J^k)$.

Assume $j-i \ge M^{\frac {3} {4}}$ (this holds whp because both $i$ and $j$ are selected uniformly at random from $[M]$), and consider the path  

\begin {displaymath}
P = (o^\emptyset, J'_1, J'_2, \ldots, J'_j, J_i, J_{i+1}, \ldots, J_M, o^*)
\end {displaymath}

which passes through $|P| \ge 3+M+M^{\frac {3} {4}}$ vertices and has weighted length $w(P)$.  We want to show that $w(P) > \max_{J \in I} \tau(J)$ whp.  By the Central Limit Theorem, (1) for any fixed $i$ and $j$, $w(P)$ is asymptotically normally distributed with mean $\mu (|P|-2)$ and standard deviation $\sigma \sqrt {(|P|-2)}$ and (2) for each $J$, $\tau(J)$ is asymptotically normally distributed with mean $\mu M$ and standard deviation $\sigma \sqrt M$.  That $w(P) > \max_{J \in I} \tau(J)$ whp follows by Chebyshev's inequality.  


\end {proof}

Lemma \ref {lem:backbone:inf} shows that as $\frac {N} {M} \rightarrow \infty$, a simple priority rule ($\pi_\infty$) almost surely generates a schedule in which no machine is idle until all the operations performed on that machine have been completed (a schedule with this property is clearly optimal).

\begin {defprioinf}
Given an $N$ by $M$ JSP instance $I$, the priority rule $\pi_\infty$ first schedules the first operation of each job (taking the jobs in order of ascending indices), then the second operation of each job, and so forth.  It is defined by $\pi_\infty(I, J^k_i) = i N + k$.  
\end {defprioinf}

\begin {lemma} \label {lem:backbone:inf}
Let $I$ be a random $N$ by $M$ JSP instance.  Then for fixed $M$, it holds whp (as $N \rightarrow \infty$) that the schedule $S = \mathcal {S} (\pi_\infty, I)$ has the property that 
\begin {displaymath}
	S(o) = S^+(\mathcal {M} (o)) \mbox { } \forall o \in ops(I) \mbox { .} 
\end {displaymath}
\end {lemma}

\begin {proof} Suppose that when executing $\pi_\infty$ we replace the line $S(o) \leftarrow \max (S^+(\mathcal {J} (o)),$  $ $ $S^+(\mathcal {M} (o)))$ (line 2(c) in the definition of a priority rule given in \S 3.3) with $S(o) \leftarrow S^+(\mathcal {M} (o))$.  If the resulting $S$ is feasible then the replacement must have had no effect.  Thus it suffices to show that the resulting $S$ is feasible whp.  Equivalently, we want to show that whp, $S(o) \ge S^+(\mathcal{J}(o))$ $\forall o \in ops(I)$ when $S$ is constructed using the modified version of line 2 (c).

Let $ops^{2+}(I) = \{J^k_i \in ops(I) : i > 1\}$ be the set of operations that are not first in their job.  It suffices to show that $S(o) - S(\mathcal{J}(o)) \ge \tau_{max}$ $\forall o \in ops^{2+}(I)$.  To this end, consider an arbitrary operation $o = J^k_i \in ops^{2+}(I)$.  Under $\pi_\infty$, the number of operations with lower priority than $o$ is $(i-1)N+(k-1)$.  The number of operations that have lower priority than $J^k_i$ \emph {and} run on machine $m(o)$ is, in expectation, equal to $\frac {1} {M} \left [ (i-1)(N-1) + (k-1) \right ]$ (where the switch from $N$ to $N-1$ is due to the fact that $o$ is the only operation in job $J^k$ that uses machine $m(o)$).  It follows that
\begin {displaymath}
\mathbb {E} [S(o)] = \frac {\mu} {M} \left [ (i-1)(N-1) + (k-1) \right ]
\end {displaymath}
so that
\begin {displaymath}
\mathbb {E} [S(o) - S(\mathcal {J} (o))] = \mathbb {E} [S(J^k_i) - S(J^k_{i-1})]= \mu \frac {N-1} {M} \mbox { .}
\end {displaymath}
In Appendix A we use a martingale tail inequality to establish the following claim. 
\\
\\
{\bf Claim 2.1. } With high probability, for all $o \in ops^{2+}(I)$ we have
\begin {displaymath}
S(o) - S(\mathcal {J} (o)) \ge \frac {1} {2} \mathbb {E} [S(o) - S(\mathcal {J} (o))] \mbox { .}
\end {displaymath}
\\
The Lemma then follows from the fact that $\frac {1} {2} \mathbb {E} [S(o) - S(\mathcal {J} (o))] > \tau_{max}$ for $N$ sufficiently large.
\end {proof}

Based on the results of computational experiments, Taillard (1994) conjectured that as $\frac {N} {M} \rightarrow \infty$ the optimal makespan is almost surely equal to the maximum machine workload.  The following corollary of Lemma \ref {lem:backbone:inf} confirms this conjecture. 

\begin {corollary} \label {optmakespan:inf}
Let $I$ be a random $N$ by $M$ JSP instance with optimal makespan $\ell_{min}(I)$.  Let $\tau(\bar m)$ $\equiv$ $\tau(\{ o \in ops(I): m(o) = \bar m \})$ denote the workload of machine $\bar m$.  Then for fixed $M$, it holds whp (as $N$$\rightarrow$$\infty$) that $\ell_{min}(I) = \max_{\bar m \in [M]} \tau(\bar m)$.
\end {corollary}


\begin {theorem} \label {backbone:inf}
Let $I$ be a random $N$ by $M$ JSP instance, and let $e$ be a randomly selected element of $E(I)$.  Then for fixed $M$, it holds whp (as $N$$\rightarrow$$\infty$) that $e \notin 1\_backbone(I)$.
\end {theorem}

\begin {proof}
Let $e = \{ J_i, J'_j \}$.  Remove both $J$ and $J'$ from $I$ to create an $N-2$ by $M$ instance $\bar I$, which comes from the same distribution as a random $N-2$ by $M$ JSP instance.  Lemma \ref {lem:backbone:inf} shows that whp there exists an optimal schedule $\bar S$ for $\bar I$ with the property described in the statement of the lemma.  

Let $\tau(\bar m)$ $\equiv$ $\tau(\{ o \in ops(\bar I): m(o) = \bar m \})$ denote the workload of machine $\bar m$ in the instance $\bar I$.  By the  Central Limit Theorem, each $\tau(\bar m)$ is asymptotically normally distributed with mean $\mu (N-2)$ and standard deviation $\sigma \sqrt {N-2}$.  It follows that whp, $|\tau(\bar m) - \tau(\bar m')| > N^{\frac {1} {4}}$ $\forall \bar m \neq \bar m'$.

Thus whp there will be only one machine still processing operations during the interval $[\ell(\bar S) - N^{\frac {1} {4}}, \ell(\bar S) ]$.  Because $\max(\tau(J), \tau(J')) \le M \tau_{max} = O(1)$, we can use this interval to construct optimal schedules containing the disjunctive arc $(J_i, J'_j)$ as well as optimal schedules containing the disjunctive arc $(J'_j, J_i)$.
\end {proof}

\section{Clustering as a Function of Makespan}

In this section we estimate the expected distance between random schedules whose makespan is within a factor $\rho$ of optimal, as a function of $\rho$ for various combinations of $N$ and $M$.  We then examine how the shape of this curve changes as a function of $\frac {N} {M}$.  More formally, if

\begin {itemize}
	\item $I$ is a random $N$ by $M$ JSP instance with optimal makespan $\ell_{min}(I)$,
	\item $\rho\_opt(I) \equiv \{S: \ell(S) \le \rho\cdot \ell_{min}(I) \}$, and
	\item $S^\rho_1$ and $S^\rho_2$ are drawn independently at random from $\rho\_opt(I)$,
\end {itemize}

we wish to compute $\mathbb {E} [ \|  S^\rho_1 - S^\rho_2  \|  ]$.

Note that the experiments of \S 4 provide an \emph {upper bound} on this quantity:

\begin {displaymath}
	\mathbb {E} \left [ \|  S^\rho_1 - S^\rho_2  \| \right ] \le M {N \choose 2} -  \mathbb {E} \left [ | \rho\_backbone |  \right ]
\end {displaymath}

but provide no lower bound (a low backbone size is not evidence that the mean distance between global optima is large).  The experiments in this section can be viewed as a test of the degree to which the upper bound provided by \S 4 is tight.

\subsection {Methodology}

We generate ``random" samples from $\rho\_opt(I)$ by running the simulated annealing algorithm of van Laarhoven et al. (1992) until it finds such a schedule.  More precisely, our procedure for sampling distances is as follows. \nocite {vanLaar92}

\begin {enumerate}

	\item Generate a random $N$ by $M$ JSP instance $I$.

	\item Using the branch and bound algorithm of Brucker et al. (1994), 
	determine the optimal makespan of $I$. \nocite {brucker94}

	\item Perform $k$ runs, $R_1, R_2, \ldots, R_k$, of the van Laarhoven et al. (1992)  
	simulated annealing algorithm.  Restart each run as many times as necessary for it to find a schedule whose makespan is optimal. \nocite {vanLaar92}

	\item For each $\rho \in \{1, 1.01, 1.02, \ldots, 1.5\}$, find the first schedule, call it {\it S$_{i}$}($\rho$), in each run {\it R$_{i}$} whose makespan is within a factor $\rho$ of optimal.  For each of the ${k \choose 2}$ pairs of runs $(R_i, R_j)$, add the distance between $S_i(\rho)$ and $S_j(\rho)$ to the sample of distances associated with $\rho$.
\end {enumerate}


We ran this procedure on random JSP instances for the same 7 combinations of $N$ and $M$ that were used in \S 4.2.  For the smallest instance sizes for each ratio (i.e., 6x6, 8x4 and 9x3 instances) we generate 100 random JSP instances and run the procedure with $k = 100$.  Setting $k = 100$ allows us to measure the variation in \emph {instance-specific} expected values.  For the other 4 combinations of $N$ and $M$, performing 10,000 simulated annealing runs is too computationally expensive, so we instead generate 1000 random JSP instances and run the procedure with $k=2$.     

Figure 5 (A), (B), and (C) plot the expected distance between random $\rho$-optimal schedules as a function of $\rho$ for each of the three values of $\frac {N} {M}$.  Figure 5 (D) shows the 0.75 and 0.25 quantiles of the 100 \emph {instance-specific} sample means for each of the three smallest instance sizes.  Examining Figure 5 (D), we see that the variation among random instances with the same $N$ and $M$ is small relative to the differences between the curves for different values of $\frac {N} {M}$.

\begin {figure} [p]
	\begin {center}
		\includegraphics [width=7.5cm,clip,trim=1cm 0.5cm 1.5cm 0cm] {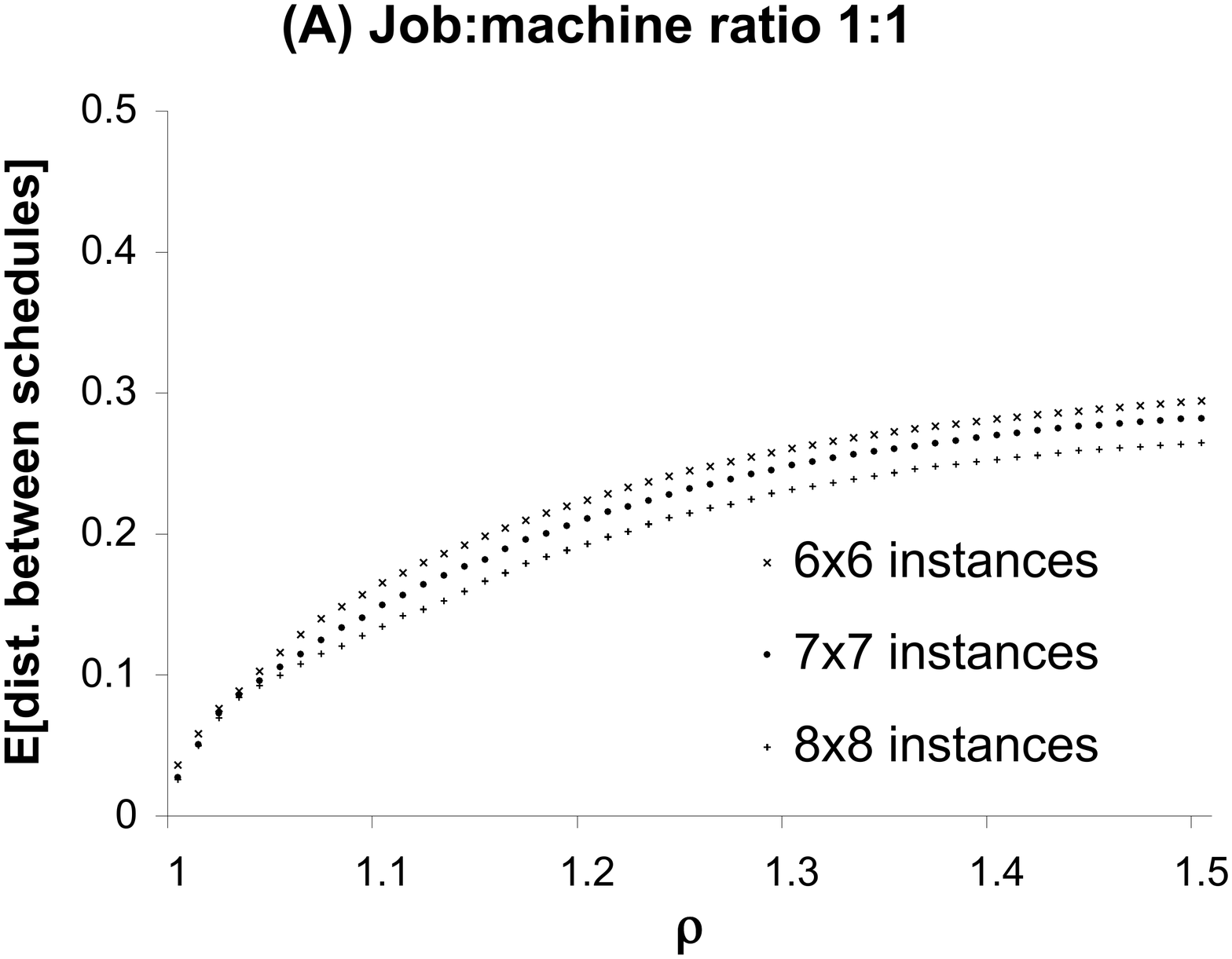} \includegraphics [width=7.5cm,clip,trim=1cm 0.5cm 1.5cm 0cm] {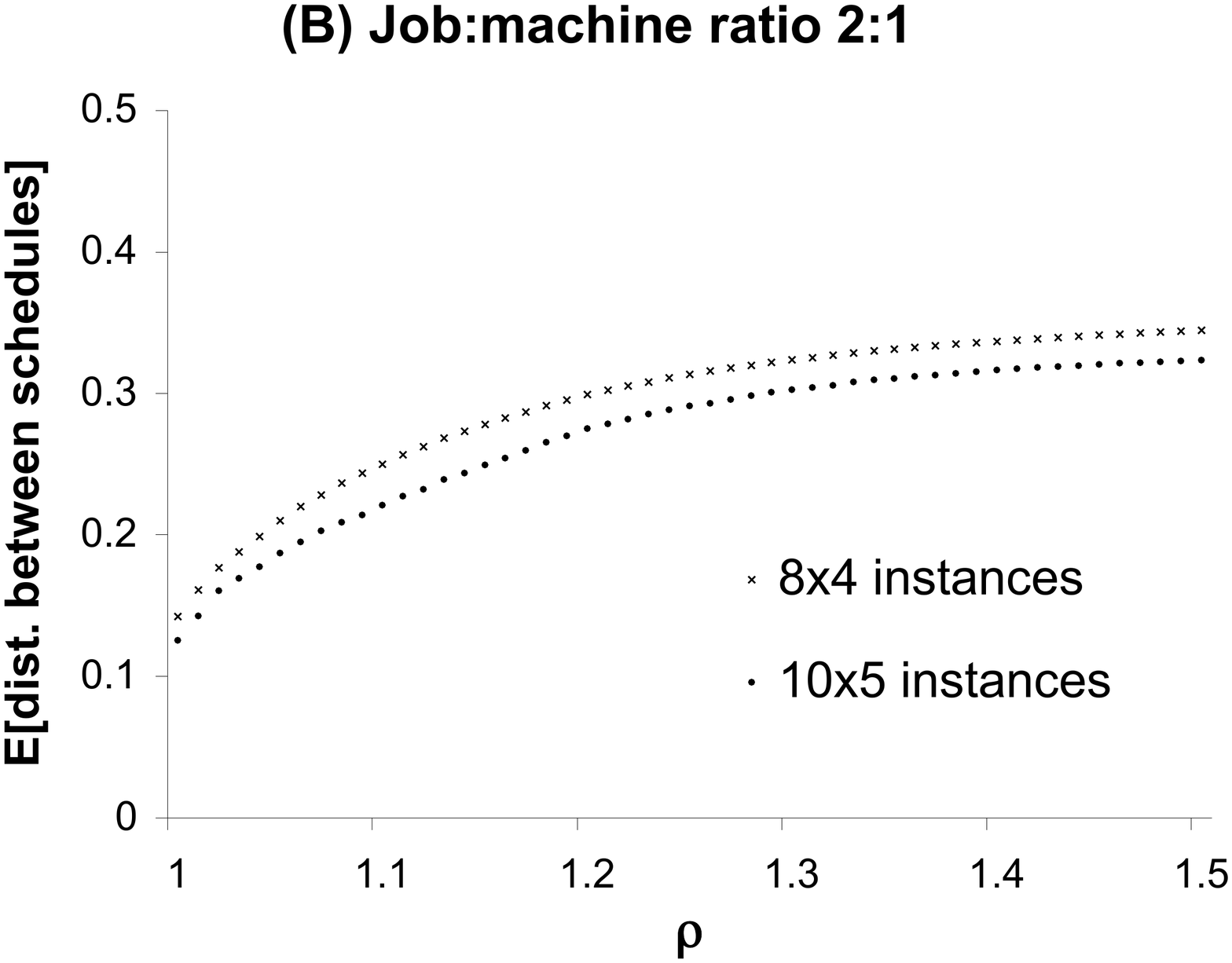} \\
		\includegraphics [width=7.5cm,clip,trim=1cm 0.5cm 1.5cm 0cm] {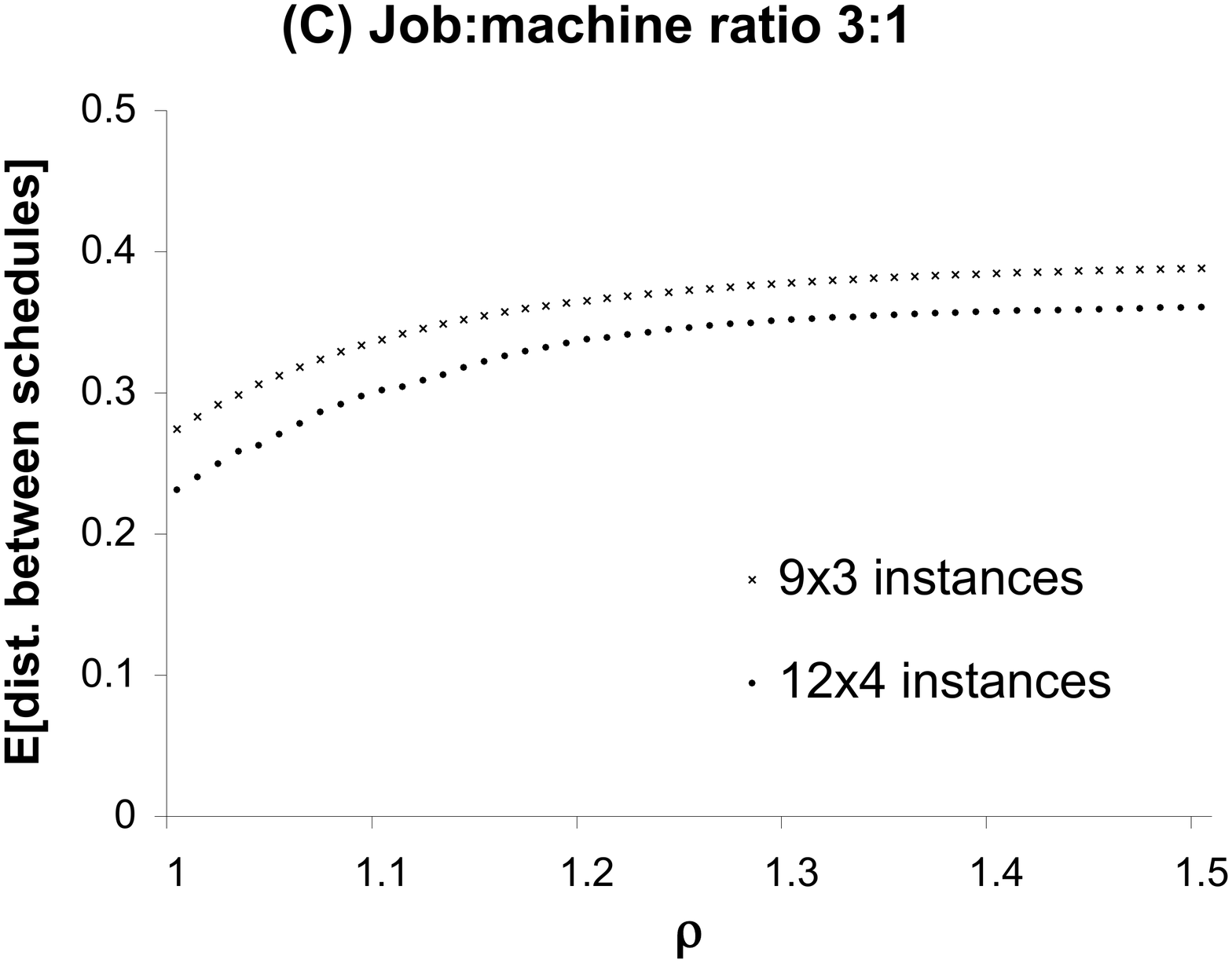} \includegraphics [width=7.5cm,clip,trim=1cm 0.5cm 1.5cm 0cm] {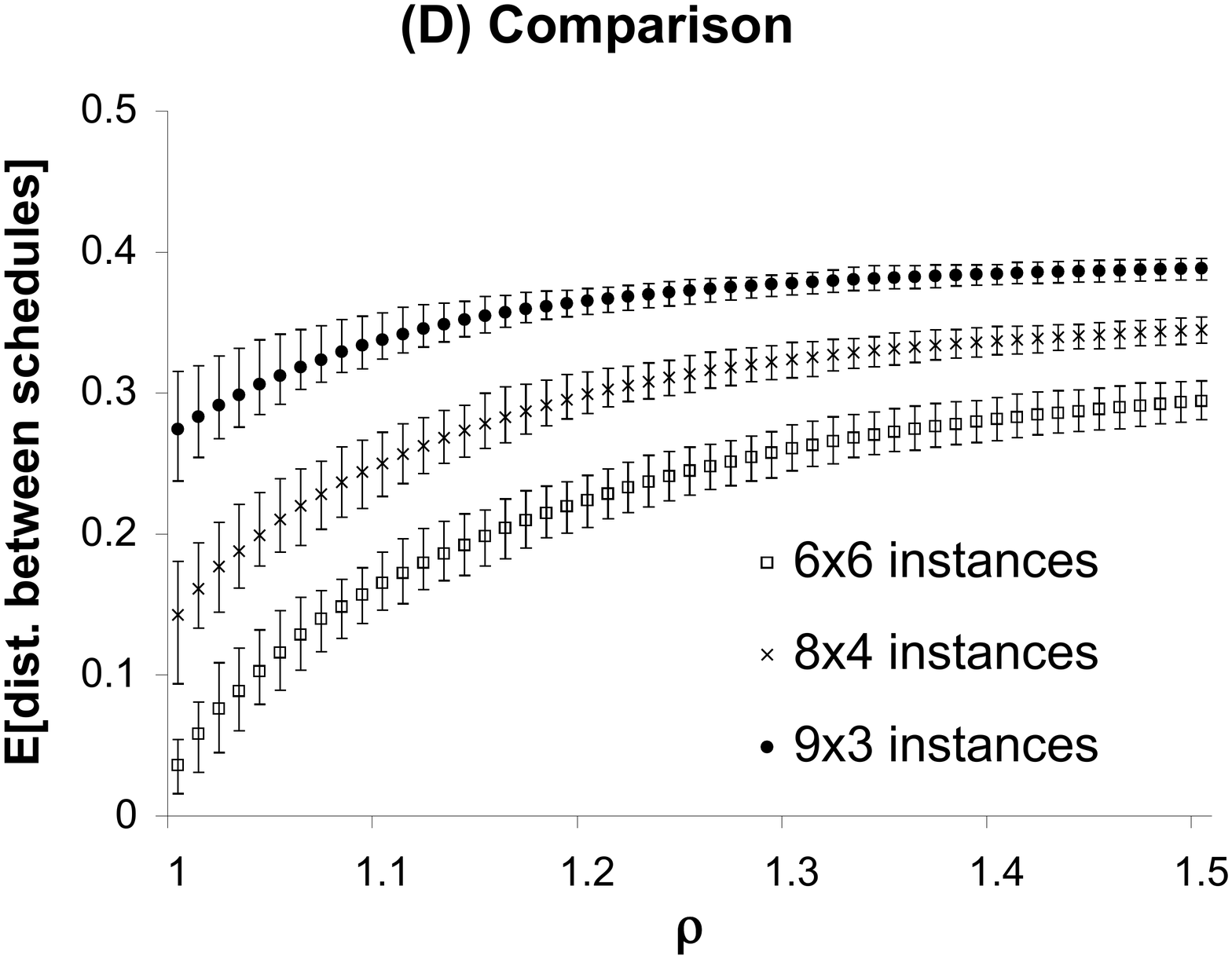}
	\end {center}
	\caption  {Expected distance between random schedules within a factor $\rho$ of optimal, as a function of $\rho$. Graphs (A), (B), and (C) depict curves for random instances with $\frac {N} {M}$ = 1, 2, and 3, respectively.  Graph (D) compares the curves depicted in (A), (B), and (C) (only the curves for the smallest instance sizes are shown in (D)).  In (D), top and bottom error bars represent 0.75 and 0.25 quantiles (respectively) of \emph {instance-specific} sample means.}
\end {figure}

\subsection {Discussion}

By examining Figure 5 we see that for any $\rho$, the expected distance between random $\rho$-optimal schedules increases as $\frac {N} {M}$ increases.  Indeed, global optima are dispersed widely throughout the search space for $\frac {N} {M} = 3$, and this is true to a lesser extent for $\frac {N} {M} = 2$.  

An immediate implication of Figure 5 is that whether or not they exhibit the two correlations that are the operational definition of a big valley, typical landscapes for JSP instances with $\frac {N} {M}=3$ cannot be expected to be big valleys in the sense of having a central cluster of optimal or near-optimal solutions.  If anything, one might posit the existence of multiple big valleys, each leading to a separate global optimum.  The next section expands upon these observations.

\section {The Big Valley}

In this section we define some formal properties of a big valley landscape, conduct experiments to determine the extent to which random JSP instances exhibit these properties as we vary $\frac {N} {M}$, and present analytical results for the limiting cases $\frac {N} {M} \rightarrow 0$ and $\frac {N} {M} \rightarrow \infty$.

Considering again the ``intuitive picture" given in Figure 1, we take the following to be necessary (though perhaps not sufficient) conditions for a function $f(x)$ to be a big valley.
\begin {enumerate}
	\item \emph {Small improving moves.} If $x$ is not a global minimum of $f$, there must exist a nearby $x'$ with $f(x') < f(x)$.
	\item \emph {Clustering of global optima.}  The maximum distance between any two global minima of $f$ is small.
\end {enumerate}
Note that there is no direct relationship between these two properties and the cost-distance correlations considered by Boese et al. (1994). \nocite {boese94}

\subsection {Formalization}

\begin {figure} \label {figbigvalleyformal}
	\begin {center}
		\includegraphics [width=14cm,clip,trim=3cm 7cm 3cm 6cm] {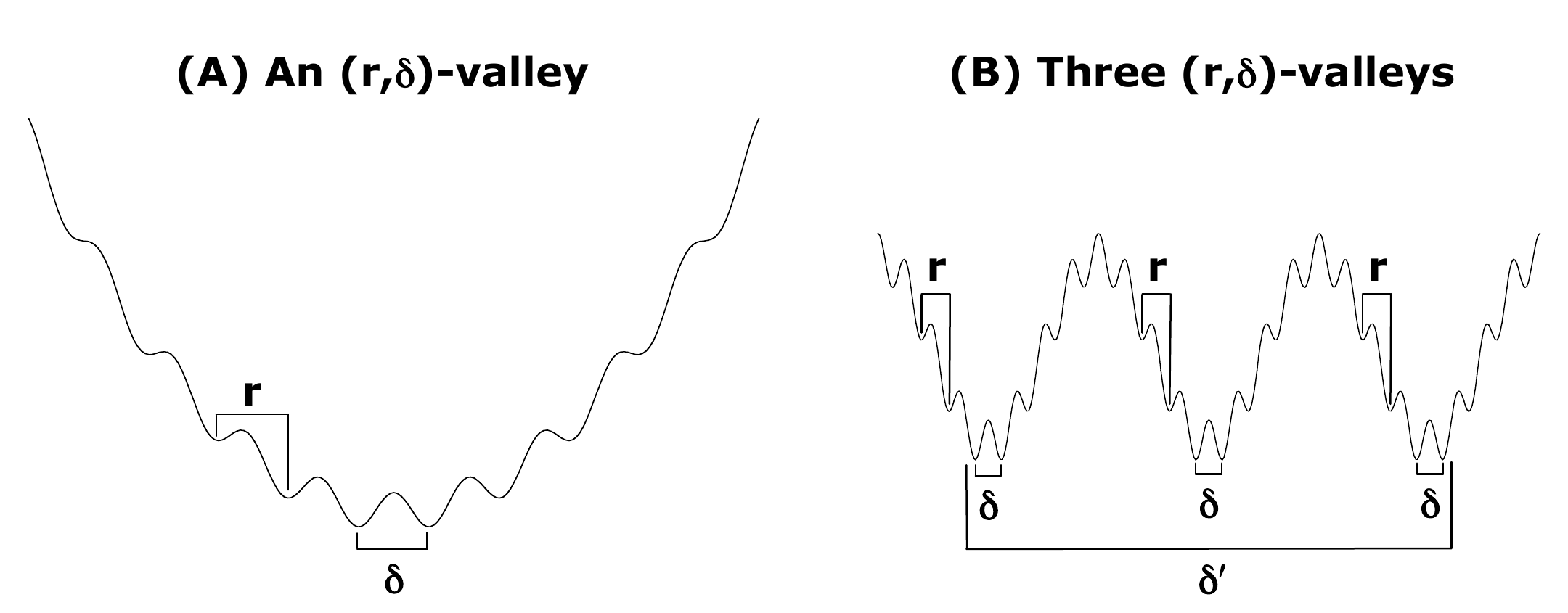}
	\end {center}
	\caption {Two landscapes comprised of $(r,\delta)$-valleys.  (A) is a single $(r,\delta)$ valley (for the values of $r$ and $\delta$ shown in the figure), while (B) can either be viewed as three distinct $(r,\delta)$ valleys or as a single $(r,\delta')$-valley.  (The values of $r$ shown in the figure are slightly larger than necessary.)}
\end {figure}

The following four definitions allow us to formalize the notion of a big valley landscape.

\begin {defnr}
Let $I$ be an arbitrary JSP instance, and let $U$ be the set of all schedules for $I$.  Let $r$ be a positive integer.  The neighborhood $\mathcal {N}_r: U \rightarrow 2^U$ is defined by
\begin {displaymath}
	\mathcal {N}_r (S) \equiv \{ S' \in U:  \| S-S' \|  \le r \} \mbox { .}
\end {displaymath}
\end {defnr}

\begin {deflopt}
Let $I$ and $U$ be as above; let $\mathcal {N}: U \rightarrow 2^U$ be an arbitrary neighborhood function; and let $S$ be a schedule for $I$.  $\mathcal {L} (S, \mathcal {N})$ is the schedule returned by the following procedure  (which finds a local optimum by performing next-descent starting from $S$ using the neighborhood $\mathcal {N}$).
\begin {enumerate}
	\item Let $\mathcal {N} (S) = \{ S_1, S_2, \ldots, S_{|\mathcal {N} (S)|} \}$ (where the elements of $\mathcal {N}(S)$ are indexed in a fixed but arbitrary manner).
	\item Find the least $i$ such that $\ell(S_i) < \ell(S)$.  If no such $i$ exists, return $S$; otherwise set $S \leftarrow S_i$ and go to 1.
\end {enumerate}
\end {deflopt}

\begin {defrdelta}
Let $I$ and $U$ be as above, and let $r$ and $\delta$ be non-negative integers.  A set $V \subseteq U$ is an \emph {$(r, \delta)$-valley} if $V$ has the following two properties.
\begin {enumerate}
	\item For any $S \in V$, the schedule $\mathcal {L} (S,\mathcal {N}_r)$ is in $V$ and is globally optimal. 
	\item For any two globally optimal schedules $S_1$ and $S_2$ that are both in $V$, $\| S_1 - S_2 \| \le \delta$.
\end {enumerate}
\end {defrdelta}

Figure 6 illustrates the definition of an $(r,\delta)$-valley.  We would say that the landscape depicted in Figure 6 (A) is a big valley, while that depicted in 6 (B) is comprised of three big valleys.

\begin {defrdeltaplandscape}
Let $I$ and $U$ be as above, and let $S$ be a random schedule for $I$.  Then $I$ has an \emph {$(r, \delta, p)$ landscape} if there exists a $V \subseteq U$ such that
\begin {enumerate}
	\item $V$ is an $(r, \delta)$-valley, and 
	\item $\mathbb {P} [S \in V] \ge p$.
\end {enumerate}
\end {defrdeltaplandscape}

Any JSP instance trivially has an $(M {N \choose 2}, M {N \choose 2}, 1)$ landscape (because if $r = M {N \choose 2}$ then $\mathcal {N}_r$ includes all possible schedules).  If a JSP instance $I$ has an $(r, M {N \choose 2}, 1)$ landscape, then a globally optimal schedule for $I$ can always be found by starting at a random schedule and applying next-descent using the neighborhood $\mathcal {N}_r$. 

We say that a JSP instance $I$ has a big valley landscape if $I$ has an $(r, \delta, p)$ landscape for small $r$ and $\delta$ in combination with $p$ near 1.  In contrast, if we have small $r$ in combination with $p$ near 1 but require large $\delta$, we say that the landscape consists of multiple big valleys.

\subsection {Neighborhood Exactness}

In this section we seek to determine the extent to which random JSP instances have the ``small improving moves" property.  We require the following definition.

\begin {defexactness}
Let $I$, $U$, and $\mathcal {N}$ be as above, and let $S$ be a random schedule for $I$.  The {\it exactness} of the neighborhood $\mathcal{N}$ on the instance $I$ is the probability that $\mathcal {L} (S, \mathcal {N})$ is a global optimum.
\end {defexactness}

If the exactness of $\mathcal {N}_r$ is $p$, then $I$ has an $(r, M {N \choose 2}, p)$ landscape (let $V$ consist of all schedules $S$ such that $\mathcal {L} (S, \mathcal {N})$ is a global optimum).  We will estimate the \emph {expected} exactness of $\mathcal {N}_r$ as a function of $r$ for various combinations of $N$ and $M$.  By examining the resulting curves, we will be able to draw conclusions about the extent to which the landscapes of a random $N$ by $M$ JSP instance typically has the ``small improving moves" property.  We can then determine how the presence or absence of this property depends on $\frac {N} {M}$.

For fixed $N$ and $M$, we compute the {\it expected} exactness of $\mathcal{N}_{r}$ for $1 \le r \le M{N \choose 2}$ by repeatedly executing the following procedure.  

\begin {enumerate}
	\item Generate a random $N$ by $M$ JSP instance $I$.
	\item Using the algorithm of Brucker et al. (1994), compute the optimal makespan of $I$.
	\item Repeat $k$ times:
	\begin {enumerate}
		\item $S \leftarrow $ a random feasible schedule, $r \leftarrow 1$, $opt \leftarrow false$.
		\item While $opt = false$ do:
		\begin {itemize}
			\item $S \leftarrow \mathcal {L} (S, \mathcal {N}_r)$.
			\item If $S$ is a global optimum, $opt \leftarrow true$.  
			\item Record the pair $(r, opt)$.
			\item $r \leftarrow r+1$.
		\end {itemize}
		\item For all $r'$ such that $r \le r' \le M{N \choose 2}$ record the pair $(r', true)$.
	\end {enumerate}
\end {enumerate}

The pairs recorded by the procedure (in step 3(c) and the third bullet point of 3 (b)) are used in the obvious way to estimate expected exactness.  Specifically, for each $r$ the estimated expected exactness of $\mathcal {N}_r$ is the fraction of pairs $(r,x)$ for which $x = true$.  

The implementation of the first bullet point in step 3 (b) deserves further discussion.  To determine $\mathcal {L} (S, \mathcal {N}_r)$, each step of next-descent must be able to determine the best schedule in $\{ S': \| S - S' \| \le r \}$.  For large $r$ it is impractical to do this by brute force.  Instead we have developed a ``radius-limited" branch and bound algorithm that, given an arbitrary center schedule $S_c$ and radius $r$, finds the schedule $\argmin_{\{ S': \| S_c - S' \| \le r \}} \ell(S')$.  Our radius-limited branch and bound algorithm uses the branching rule of Balas (1969) combined with the lower bounds and branch ordering heuristic of Brucker et al. (1994). \nocite {balas69,brucker94}   

\subsection {Results}

We use three combinations of $N$ and $M$ with $\frac {N} {M} = \frac {1} {5}$ (3x15, 4x20, and 5x25 instances), three combinations with $\frac {N} {M} = 1$ (6x6, 7x7, and 8x8 instances) and two combinations with $\frac {N} {M} = 5$ (15x3 and 20x4 instances).  For the smallest instance sizes for each ratio (i.e., 3x15, 6x6, and 15x3 instances) we generate 100 random JSP instances and run the above procedure with $k = 100$.  Otherwise, we generate 1000 random JSP instances and run the procedure with $k = 1$.

Figure 7 (A), (B), and (C) plot expected exactness as a function of neighborhood radius (normalized by the number of disjunctive edges) for each of these three values of $\frac {N} {M}$.  Figure 7 (D) shows the 0.75 and 0.25 quantiles of the 100 \emph {instance-specific} sample means for each of the three smallest instance sizes.

\begin {figure} [p]
	\begin {center}
		\includegraphics [width=7.5cm,clip,trim=1cm 0.5cm 1.5cm 0cm] {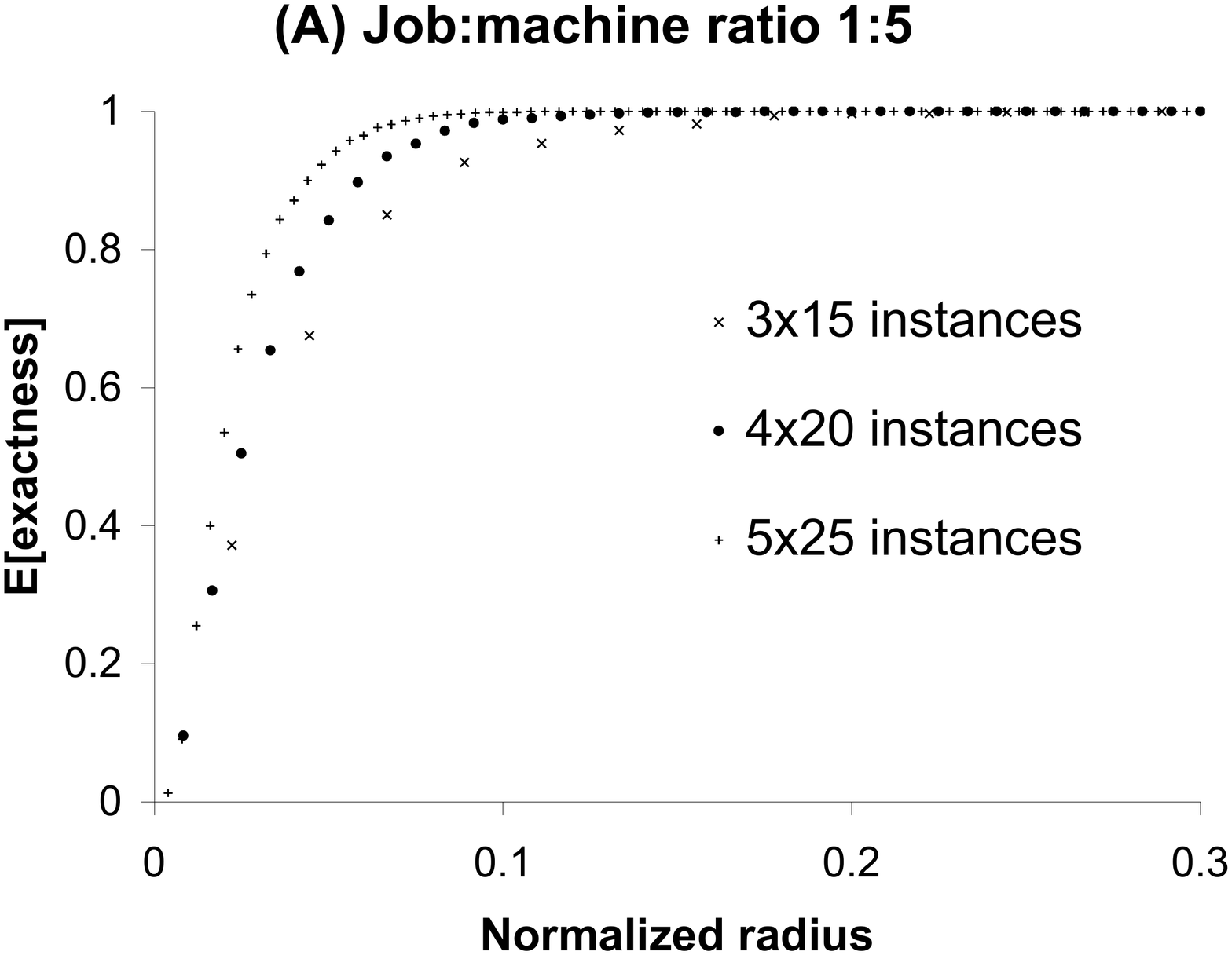} \includegraphics [width=7.5cm,clip,trim=1cm 0.5cm 1.5cm 0cm] {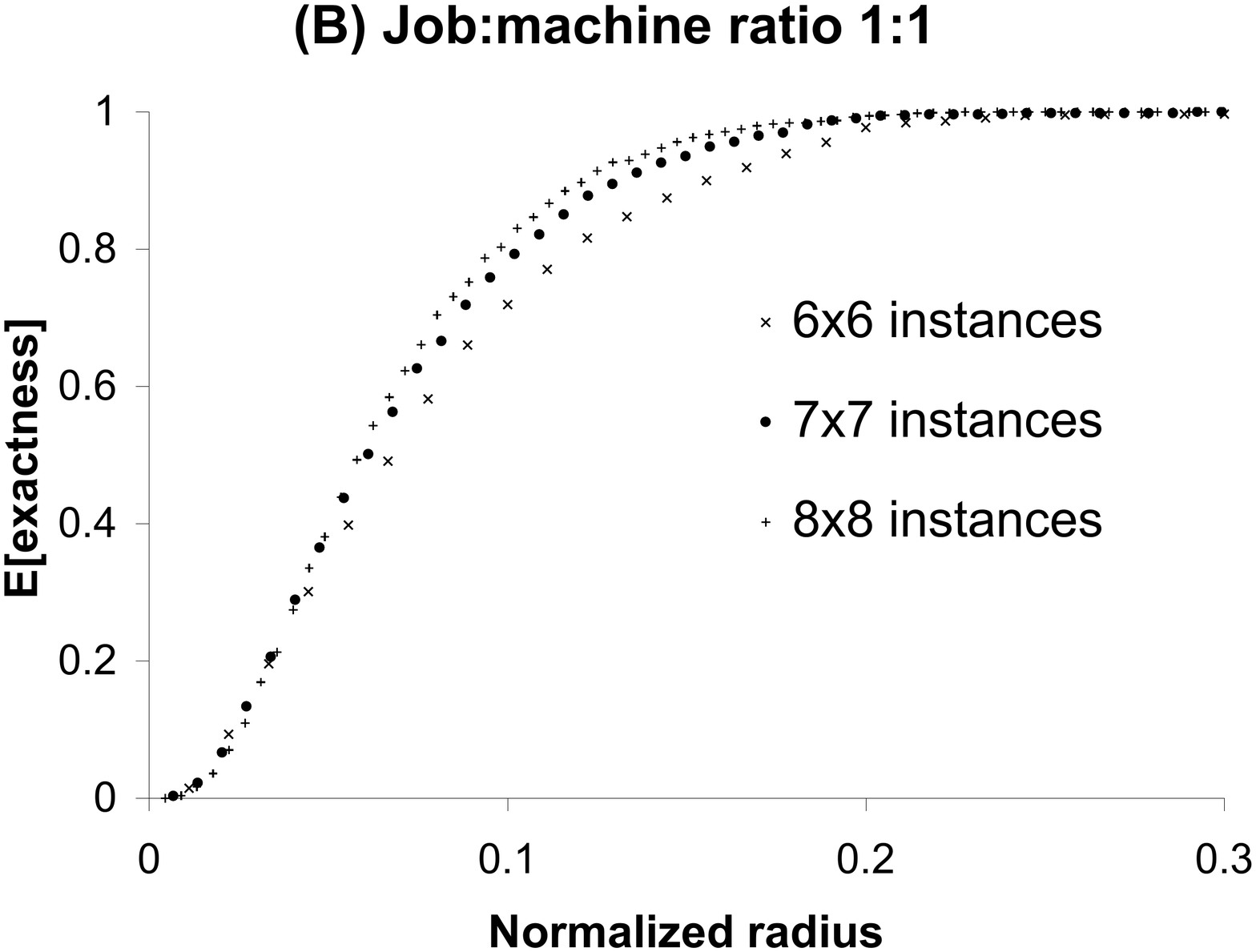} \\
		\includegraphics [width=7.5cm,clip,trim=1cm 0.5cm 1.5cm 0cm] {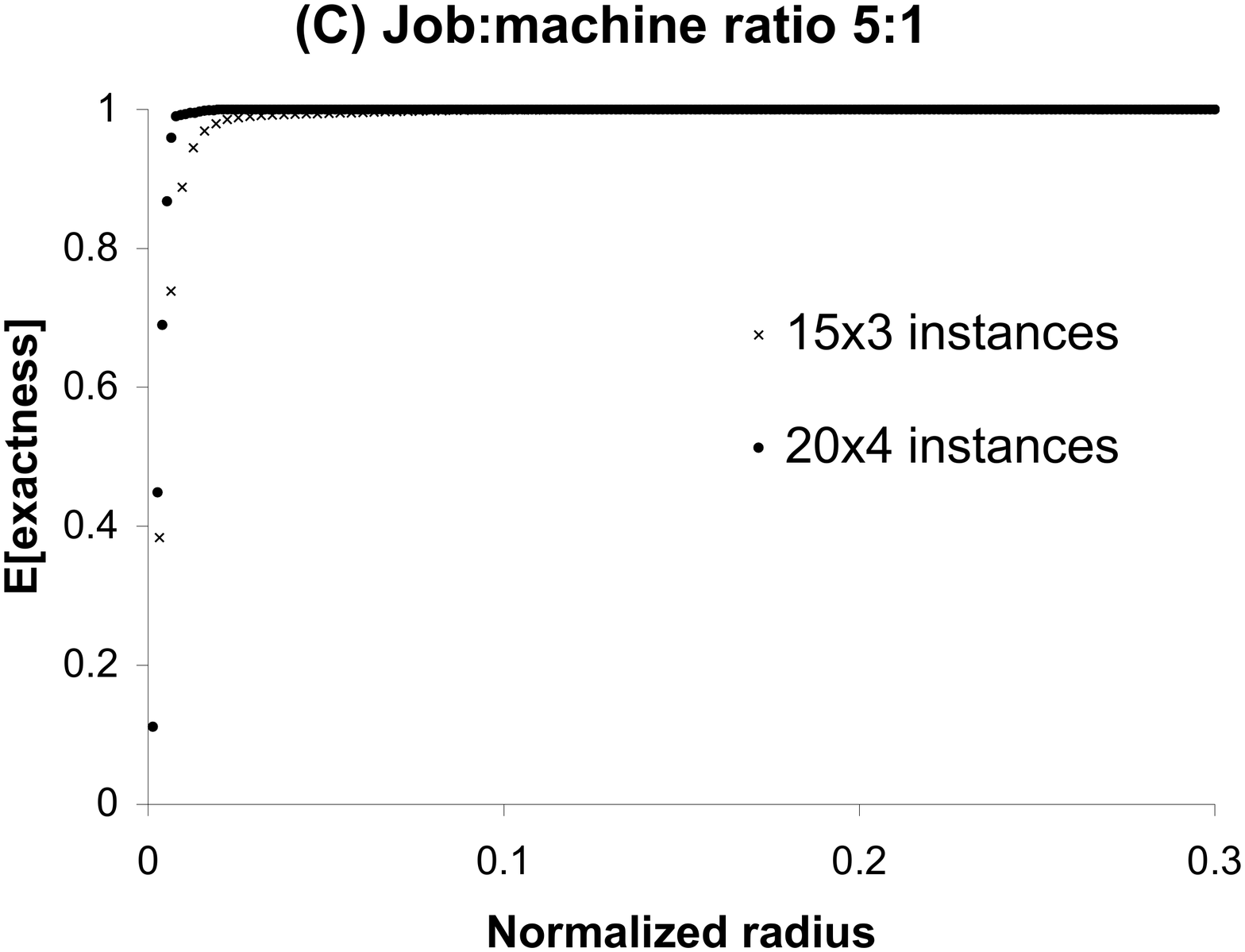} \includegraphics [width=7.5cm,clip,trim=1cm 0.5cm 1.5cm 0cm] {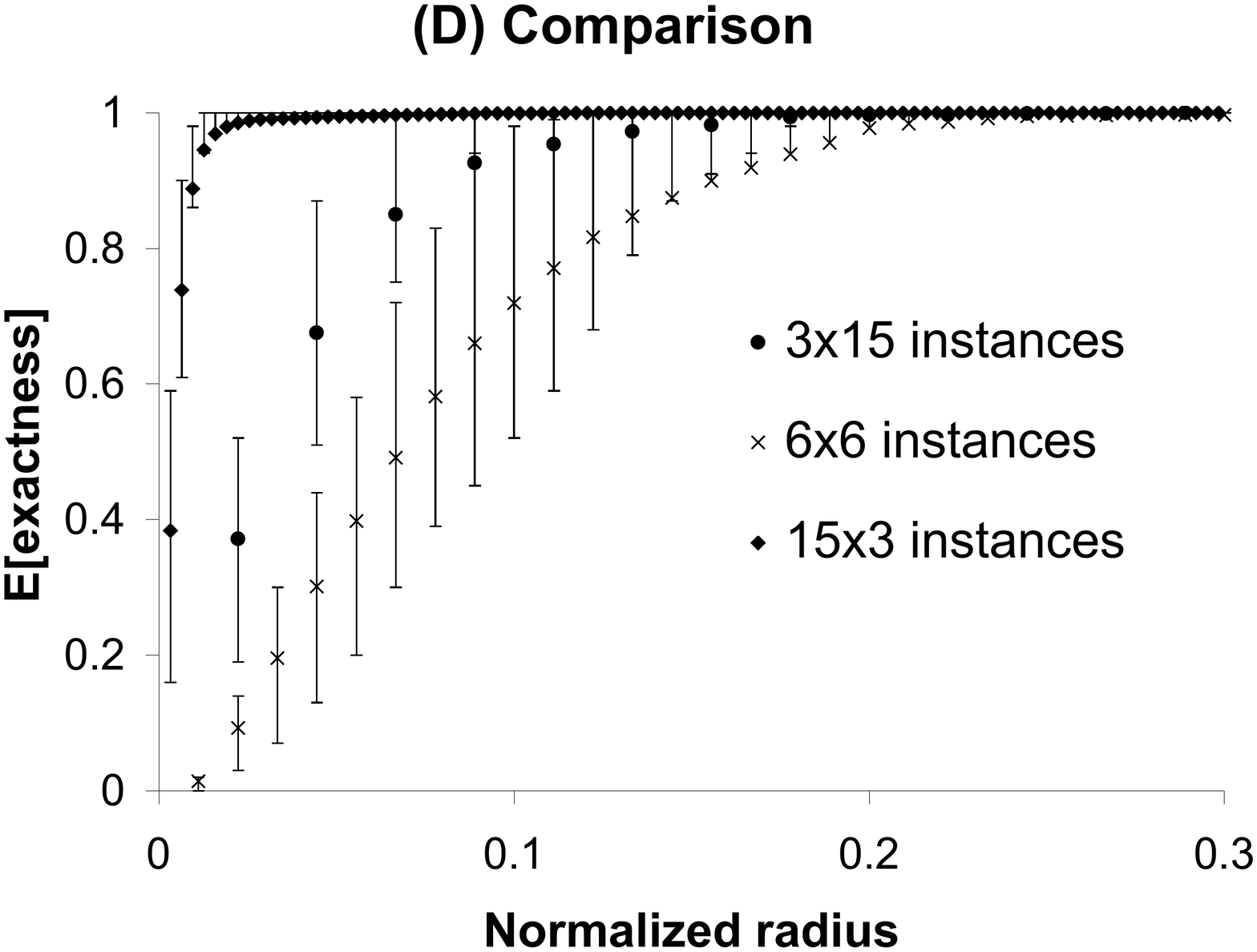}
	\end {center}
	\caption  {Expected exactness of $\mathcal {N}_r$ as a function of the (normalized) neighborhood radius $r$. Graphs (A), (B), and (C) depict curves for random instances with $\frac {N} {M}$ =  $\frac {1} {5}$, 1, and 5, respectively.  Graph (D) compares the curves depicted in (A), (B), and (C) (only the curves for the largest instances are shown in (D)).  In (D), top and bottom error bars represent 0.75 and 0.25 quantiles (respectively) of \emph {instance-specific} exactness.}
\end {figure}

\subsection {Discussion}

Examining Figure 7, we see that for any normalized neighborhood radius, the neighborhood exactness is lowest for instances with $\frac {N} {M} = 1$ and higher for the two more extreme ratios ($\frac {N} {M} = \frac {1} {5}$ and $\frac {N} {M} = 5$).  If we view neighborhood exactness as measuring the ``smoothness" of a landscape, the data suggest that typical JSP landscapes are least smooth at some intermediate value of $\frac {N} {M}$, but become more smooth as $\frac {N} {M} \rightarrow 0$ or $\frac {N} {M} \rightarrow \infty$.  This in itself suggests an easy-hard-easy pattern of typical-case instance difficulty in the JSP, a phenomenon explored more fully in the next section.

Using the methodology of \S\S 4-5, we found that the expected proportions of backbone edges for 3x15, 4x20, and 5x25 instances are 0.94, 0.93, and 0.92, respectively, while the expected distance between global optima was 0.02 in all three cases.  In contrast, the expected proportions of backbone edges for 15x3 and 20x4 instances are near-zero, while the expected distances between global optima are 0.33 and 0.28, respectively.  We conclude that landcapes of random $N$ by $M$ JSP instances typically have the ``clustering of global optima" property for $\frac {N} {M} = \frac {1} {5}$ but not for $\frac {N} {M} = 5$.  However, Figure 7 suggests that the ``small improving moves" property is present for both $\frac {N} {M} = \frac {1} {5}$ and $\frac {N} {M} = 5$.  Accordingly, we would say that typical landscapes for $\frac {N} {M} = \frac {1} {5}$ are big valleys, while for $\frac {N} {M} = 5$ the landscape is comprised of many big valleys rather than just one.

The data from \S\S4-5 show that for $\frac {N} {M} = 1$, typical landscapes have the ``clustering of global optima" property.  Examining Figure 7 (B), we see that we are able to descend from a random schedule to a globally optimal schedule with probability $\frac {1} {2}$ when the (normalized) neighborhood radius is about 6\%.  For this reason, we think of the landscapes of random JSP instances with $\frac {N} {M} = 1$ as having the ``small improving moves" property to some extent.  This, in combination with the curve in Figure 5 (A) (which shows  expected distance between random $\rho$-optimal schedules as a function of $\rho$) leads us to say that typical landscapes of random JSP instances with $\frac {N} {M} = 1$ can still be roughly described as big valleys.  However, the valley is much rougher (meaning that larger steps are required to move from a random schedule to a global optimum via a sequence of improving moves) than for the more extreme values of $\frac {N} {M}$.

Table 1 summarizes the empirical findings just discussed.
\begin {center}
\begin {tabular} {c}
Table 1. Landscape attributes for three values of $\frac {N} {M}$. \\
\begin {tabular} {| l | l | l | l |}
	\hline 
	\bf {$\frac {N} {M}$}	& \parbox {3 cm} {\bf {Clustering of global optima?}} & \parbox {4 cm} {\bf {Small improving moves?}} & \bf {Description} \\[6pt] \hline 
	$\frac {1} {5}$	& Yes	& Yes & Big valley \\[6pt] \hline
	$1$			& Yes	& Somewhat & (Rough) big valley \\[6pt] \hline
	$5$			& No		& Yes	&	Multiple big valleys \\[6pt] \hline
\end {tabular} \\
\end {tabular}
\end {center}

\subsection {Analysis}

We first establish the behavior of the curves depicted in Figure 7 in the limiting cases $\frac {N} {M} \rightarrow 0$ and $\frac {N} {M} \rightarrow \infty$.  We then use these results to characterize the landscapes of random JSP instances using the $(r, \delta, p)$ notation introduced in \S 6.1.

The following two lemmas show that as $\frac {N} {M} \rightarrow 0$ (resp. $\frac {N} {M} \rightarrow \infty$), a random schedule will almost surely be ``close" to an optimal schedule.  The proofs are given in Appendix A.

\begin {lemma} \label {closeness:0}
	Let $I$ be a random $N$ by $M$ JSP instance, and let $S$ be a random schedule for $I$.  Let $\hat S$ be an optimal schedule for $I$ such that $\| S - \hat S \|$ is minimal.  Let $f(M)$ be any unbounded, increasing function of $M$.  Then for fixed $N$, it holds whp (as $M \rightarrow$ $\infty$) that $\| S - \hat S \| < f(M)$.
\end {lemma}

\begin {lemma} \label {closeness:inf}
	Let $I$ be a random $N$ by $M$ JSP instance, let $S$ be a random schedule for $I$, and let $\hat S$ be an optimal schedule for $I$ such that $\| S - \hat S\|$ is minimal.  Then for fixed $M$ and $\epsilon > 0$, it holds whp (as $N \rightarrow$ $\infty$) that $\| S - \hat S\| < N^{1+\epsilon}$.
\end {lemma}

The following are immediate corollaries of Lemmas \ref {closeness:0} and \ref {closeness:inf}.


\begin {corollary} \label {exactness:0}
	For fixed $N$, the expected exactness of $\mathcal {N}_{f(M)}$ approaches 1 as $M \rightarrow \infty$, where $f(M)$ is any unbounded, increasing function of $M$.
\end {corollary}


\begin {corollary} \label {exactness:inf}
	For fixed $M$ and $\epsilon > 0$, the expected exactness of $\mathcal {N}_{N^{1+\epsilon}}$ approaches 1 as $N \rightarrow \infty$.
\end {corollary}

Because the total number of disjunctive edges is $M {N \choose 2}$, these two corollaries imply that as $\frac {N} {M} \rightarrow 0$ (resp. $\frac {N} {M} \rightarrow \infty$), the curve depicted in Figure 7 approaches a horizontal line at a height of 1.

Using Lemmas \ref {closeness:0} and \ref {closeness:inf}, Theorems \ref {bigvalley:0} and \ref {bigvalley:inf} characterize the landscape of random JSP instances using the $(r, \delta, p)$ notation of \S 6.1. Before presenting these theorems, a slight disclaimer is in order.  Lemmas \ref {closeness:0} and \ref{closeness:inf} (the proofs of which are fairly involved) indicate that in the extreme cases $\frac {N} {M} \rightarrow 0$ and $\frac {N} {M} \rightarrow \infty$ we can jump from a random schedule to a globally optimal schedule via a single small move.  We strongly believe that in these cases it is also possible to go from a random schedule to a global optimum by a sequence of many (smaller) improving moves, although proving this seems difficult.  Nevertheless, it should be understood that our theoretical results do not strictly imply the existence of landscapes like those depicted in Figure 6 (where for most starting points there is a sequence of two or more small improving moves leading to a global optimum).

Theorem \ref {bigvalley:0} shows that as $\frac {N} {M} \rightarrow 0$, a random JSP instance almost certainly has an $(r, \delta, p)$ landscape where $r$ grows arbitrarily slowly as a function of $M$, $\delta$ is $o({M {N \choose 2}})$, and $p$ is arbitrarily close to 1.  In other words, as $\frac {N} {M} \rightarrow 0$ the landscape has both the ``small improving move(s)" property and the ``clustering of global optima" property.  In contrast, Theorem \ref {bigvalley:inf} shows that as $\frac {N} {M} \rightarrow \infty$, a random JSP instance almost surely does not have an $(r, \delta, p)$ landscape unless $\delta$ is $\Omega(N^2)$.  Instead, the landscape contains $\Omega(N!)$ $(r, 1)$-valleys, where $r$ is $o({M {N \choose 2}})$.  Thus, as $\frac {N} {M} \rightarrow \infty$, the landscape has the ``small improving move(s)" property but not the ``clustering of global optima" property.  These analytical results confirm the trend suggested by Figure 7 and discussed in \S 6.4.


\begin {theorem} \label {bigvalley:0}
	Let $I$ be a random $N$ by $M$ JSP instance.  Let $f(M)$ be any unbounded, increasing function of $M$.  For fixed $N$ and $\epsilon > 0$, it holds whp (as $M \rightarrow \infty$) that $I$ has a $(r, \delta, p)$ landscape for $r=f(M)$, $\delta = \epsilon M {N \choose 2}$ and $p = 1-\epsilon$.
\end {theorem}

\begin {proof}
Let $V$ be the set of all schedules $S$ such that $\mathcal {L} (S, \mathcal {N}_{r})$ is a global optimum.  It follows by Corollary \ref {exactness:0} that whp, the exactness of $I$ on $r$ is at least $p$, which means $S \in V$ with probability at least $p$.  It remains to show that $V$ is an $(r, \delta)$-valley whp.  Part 1 of the definition of an $(r, \delta)$-valley is satisfied by the definition of $V$.  Part 2 follows from Theorem \ref {backbone:0}.
\end {proof}


\begin {theorem} \label {bigvalley:inf}
	Let $I$ be a random $N$ by $M$ JSP instance, and let $S$ be a random schedule for $I$.  There exists a set $V(I) = \cup_{i=1}^n V_i$ of schedules for $I$ such that for fixed $M$ and $\epsilon > 0$, $V$ has the following properties whp:
	\begin {enumerate}
		\item $S \in V$;
		\item $V_i$ is an $(r, \delta)$-valley with $r=N^{1+\epsilon}$ and $\delta=1$ $\forall i \in [n]$;
		\item $n > N! (1-\epsilon)$; and
		\item $\max_{\{S_1, S_2\} \subseteq V} \| S_1 - S_2 \| >  \Omega(N^2)$.
	\end {enumerate}
\end {theorem}

\begin {proof}
Let $\{\hat S_1, \hat S_2, \ldots, \hat S_n\}$ be the set of globally optimal schedules for $I$, and define $V_i \equiv \{ S:  \mathcal {L} (S, \mathcal {N}^{1+\epsilon}) = \hat S_i \}$.  Property 1 holds whp by Lemma \ref {closeness:inf}.  Property 2 holds by definition of $V_i$.  

The fact that property 3 holds whp is a consequence of Lemma \ref {lem:backbone:inf}.  Recall that Lemma \ref {lem:backbone:inf} showed that as $\frac {N} {M} \rightarrow \infty$, the priority rule $\pi_\infty$ generates an optimal schedule whp, where $\pi_\infty(I,J^k_i) = iN + k$.  Because the indices assigned to the jobs are arbitrary, Lemma \ref {lem:backbone:inf} also applies to the priority rule $\pi^{\phi} (I,J^k_i) = iN + \phi(k)$, where $\phi$ is any permutation of $[N]$.  There are $N!$ possible choices of $\phi$.  Let $f$ be the number of choices that fail to yield a globally optimal schedule.  Property 3 can only fail to hold if $f \ge \epsilon N!$.  But by Lemma \ref {lem:backbone:0}, $\mathbb {E} [f]$ is $o(1) N!$; hence $f < \epsilon N!$ whp by Markov's inequality.

To establish property 4, choose permutations $\phi_1$ and $\phi_2$ that list the elements of $[N]$ in reverse order (i.e., $\phi_1(i) = \phi_2(N-i)$ $\forall i \in [N]$).  By Lemma 2, the schedules $S_1 = \mathcal {S} (\pi^{\phi_1}, I)$ and $S_2 = \mathcal {S} (\pi^{\phi_2}, I)$ are both globally optimal whp.  But for any disjunctive edge $e = \{ J_1, J_1' \}$ we must have $\vec e(S_1) \neq \vec e(S_2)$, hence $\| S_1 - S_2 \| \ge |\{ \{ J, J' \} \subseteq I: m(J_1) = m(J_1')  \}| \ge  {{N  M^{-1}} \choose 2} = \Omega(N^2)$, where we obtain the expression ${{N  M^{-1}} \choose 2}$ using the pigeonhole principle.
\end {proof}

\section {Quality of Random Schedules}

\subsection {Methodology}

In this section we examine how the quality of randomly generated schedules changes as a function of the job:machine ratio.  Specifically, for various combinations of $N$ and $M$, we estimate the expected value of the following four quantities:

\renewcommand{\labelenumi}{(\Alph{enumi})}
\begin {enumerate}
	\item the makespan of a random schedule,
	\item the makespan of a locally optimal schedule obtained by starting at a random schedule and applying next-descent using the $\mathcal {N}_1$ move operator,
	\item the makespan of an optimal schedule, and
	\item the lower bound on the makespan of an optimal schedule given by the maximum of the maximum job duration and the maximum machine workload: 
	\begin {displaymath}
		\max \left ( \max_{J \in I} \tau(J), \max_{\bar m \in [M]} \sum_{o \in ops(I): m(o) = \bar m} \tau(o) \right ) \mbox { .}
	\end {displaymath}
\end {enumerate}
\renewcommand{\labelenumi}{\emph{\arabic{enumi}.}}

The values of $\frac {N} {M}$ considered in our experiments are those in the set $R = \{$$\frac {1} {7}$, $\frac {1} {6}$, $\frac {1} {5}$, $\frac {1} {4}$, $\frac {1} {3}$, $\frac {1} {2}$, $\frac {2} {3}$, $1$, $\frac {3} {2}$, $2$, $3$, $4$, $5$, $6$, $7$ $\}$.  We consider all combinations of $N$ and $M$ in the set $S \equiv \bigcup_{r \in R} S_r$, where $S_r \equiv \{ (N,M): \frac {N} {M} = r, \min(N,M) \ge 2, \max(N,M) \ge 6, NM < 1000 \}$.  For each $(N,M) \in S$, we estimate the expected value of (A) (resp. (B)) by generating 100 random $N$ by $M$ JSP instances and, for each instance, generating 100 random schedules (resp. local optima).  We estimate (D) by generating 1000 random JSP instances for each $(N,M) \in S$.  For some combinations $(N,M) \in S_{small} \subseteq S$, it was also practical to compute quantity (C).  Let $n_r = |S_{small} \cap S_r|$ be the number of combinations $(N,M)$ with $\frac {N} {M} = r$ for which we computed (C).  We chose $S_{small}$ so that $n_r \ge 4$ for $r \neq \frac {3} {2}$ while $n_{\frac {3} {2}} = 3$.  For each $(N,M) \in S_{small}$, we estimate (C) using 1000 random JSP instances.

\subsection {Results}

Figure 8 plots the mean values of (A), (B), and (C), respectively, against the mean value of (D), for various combinations of $N$ and $M$.  The data points for each combination of $N$ and $M$ are assigned a symbol based on the value of $\frac {N} {M}$.  Top and bottom error bars represent 0.75 and 0.25 quantiles (respectively) of \emph {instance-specific} sample means.  Note that the width of these error bars is small relative to the differences between the curves for different values of $\frac {N} {M}$.

Examining Figure 8, we see that the set of data points for each value of $\frac {N} {M}$ are approximately (though not exactly) collinear.  Furthermore, in all three graphs the slope of the line formed by the data points with $\frac {N} {M} = r$ is maximized when $r=1$, and decreases as $r$ gets further away from 1 (see also Figure 9 (A)). 

To further investigate this trend, we performed least squares linear regression on the set of data points for each value of $\frac {N} {M}$.  The slopes of the resulting lines are shown as a function of $\frac {N} {M}$ in Figure 9 (A).

\begin {figure} [p]
	\begin {center}
		\includegraphics [width=8.5cm,clip,trim=1cm 0.5cm 0cm 1cm] {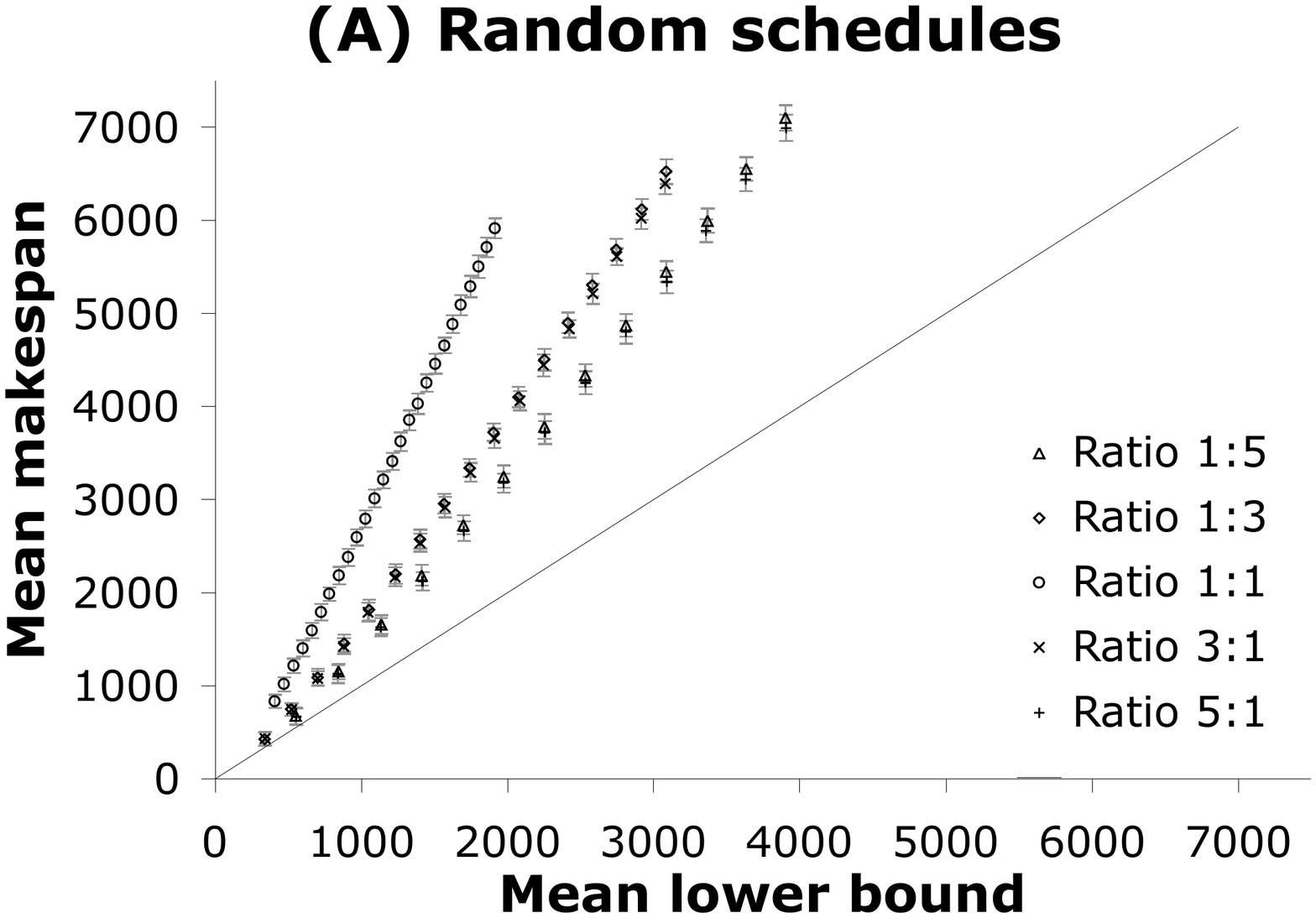} \\
		\includegraphics [width=8.5cm,clip,trim=1cm 0.5cm 0cm 1cm] {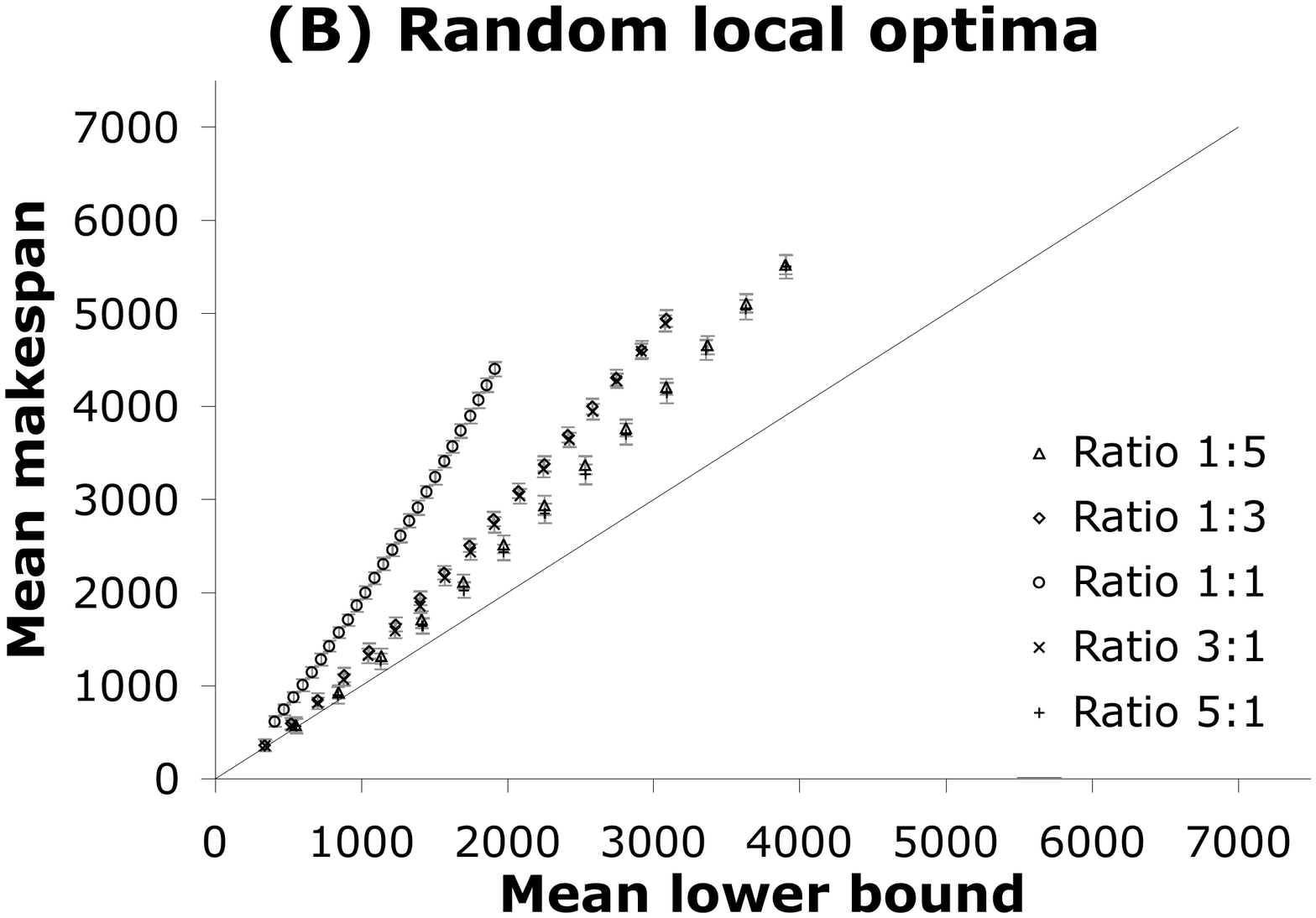} \\
		\includegraphics [width=8.5cm,clip,trim=1cm 0.5cm 0cm 1cm] {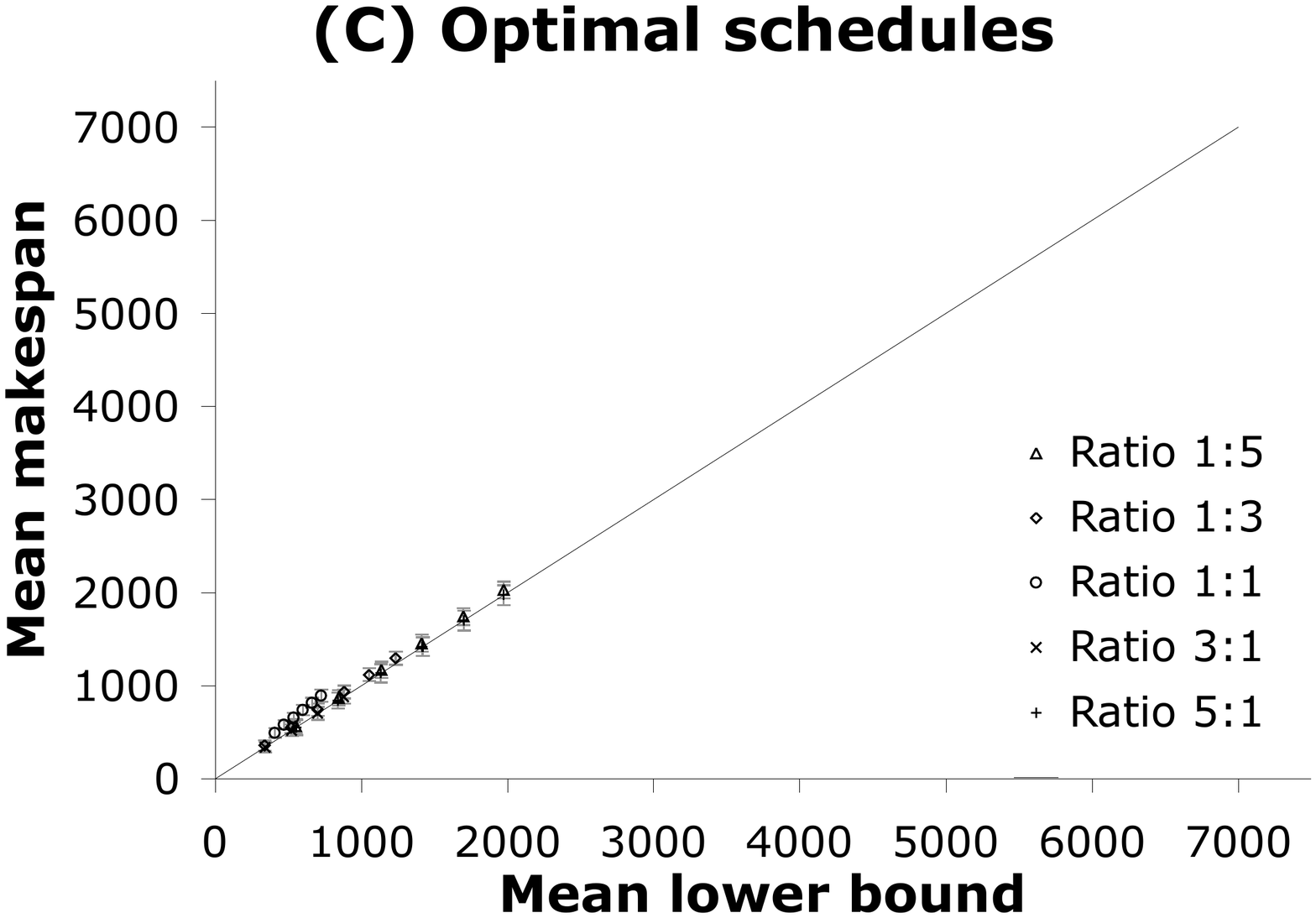}
	\end {center}
	\caption  {Expected makespan of (A) random schedules, (B) random local optima, and (C) optimal schedules vs. expected lower bound, for various combinations of $N$ and $M$ (grouped by symbol according to $\frac {N} {M}$).  Top and bottom error bars represent 0.75 and 0.25 quantiles (respectively) of \emph {instance-specific} sample means.}
\end {figure}

\begin {figure} [p]
	\begin {center}
		\includegraphics [width=12cm,clip,trim=1cm 0cm 0cm 0cm] {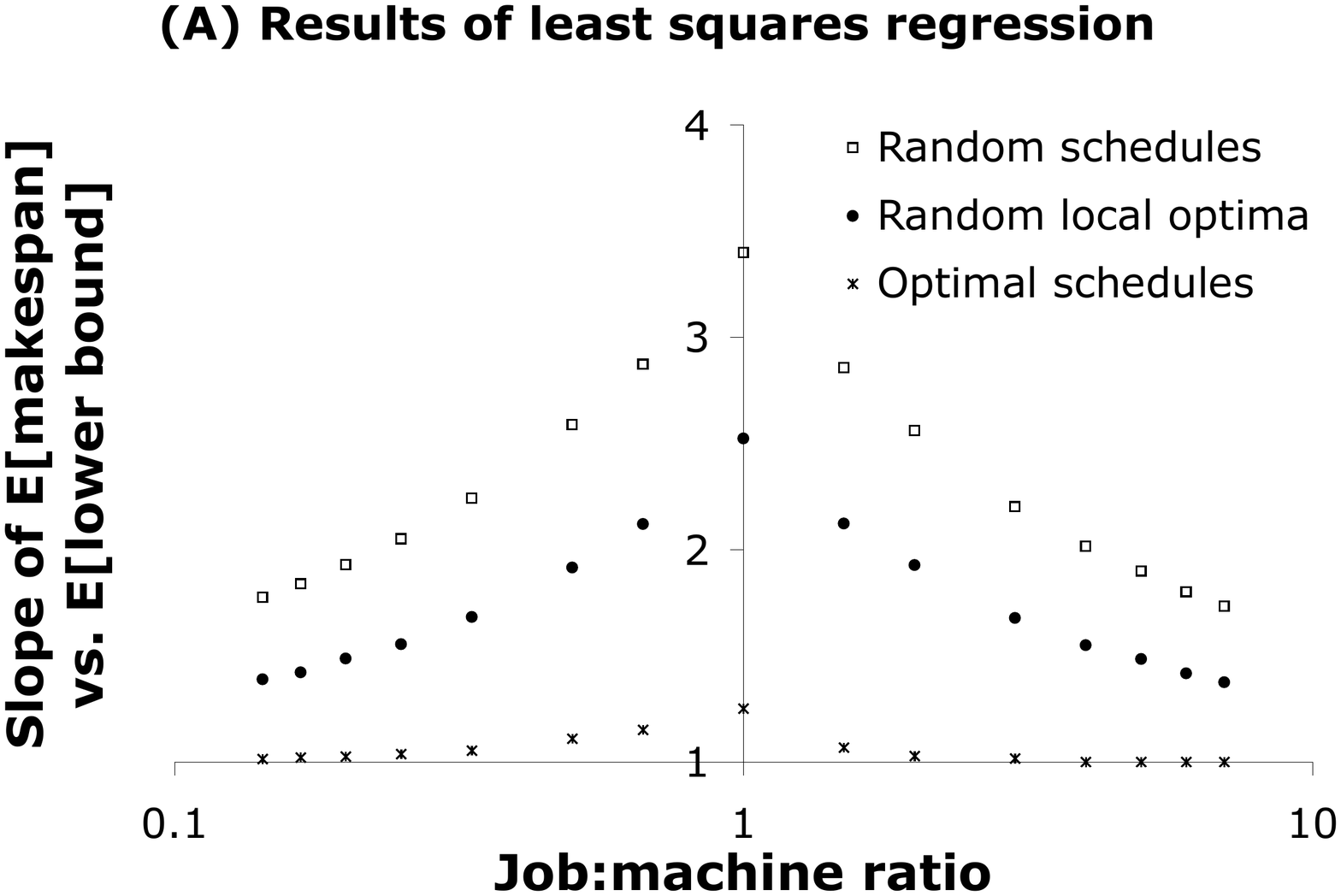}
		\includegraphics [width=12cm,clip,trim=1cm 1cm 0cm 2cm] {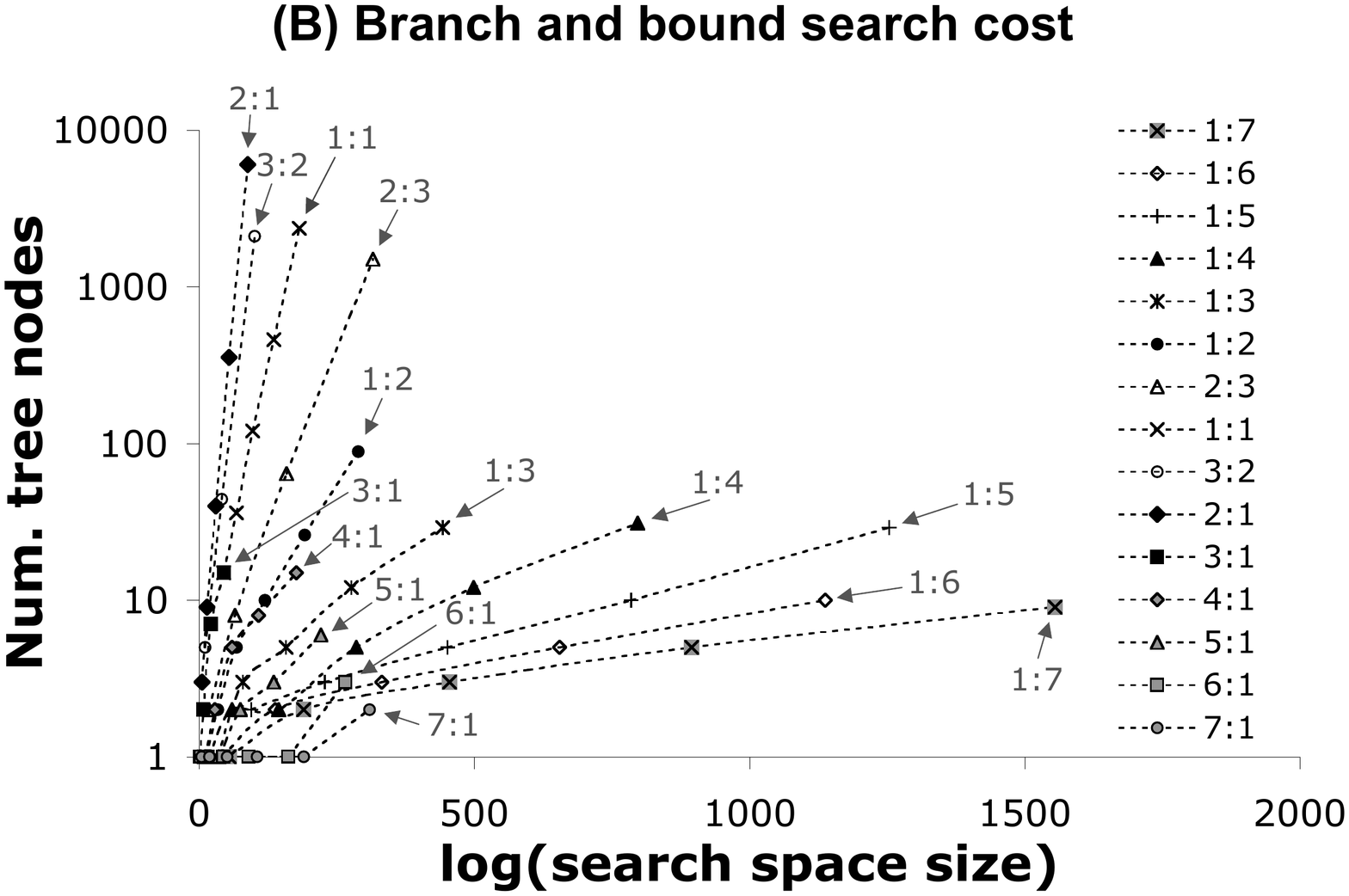}
	\end {center}
	\caption  {(A) graphs the slope of the least squares fits to the data in Figure 8 (A), (B), and (C) as a function of $\frac {N} {M}$ (includes values of $\frac {N} {M}$ not depicted in Figure 8).  (B) graphs the number of search tree nodes ($90^{th}$ percentile) used by the branch and bound algorithm of Brucker et al. (1994) to find an optimal schedule.}
\end {figure}

From examination of Figure 9 (A), it is apparent that

\begin {itemize}
	\item as the value of $\frac {N} {M}$ becomes more extreme (i.e., approaches either 0 or $\infty$), the expected makespan of random schedules (resp. random local optima) comes closer to the expected value of the lower bound on makespan; and

	\item the difference between the expected makespan of random schedules (resp. random local optima) and the expected value of the lower bound on makespan is maximized at a value of $\frac {N} {M} \approx 1$.
\end {itemize}

The first of these two observations suggests that as $\frac {N} {M}$ approaches either $0$ or $\infty$, a random schedule is almost certainly near-optimal.  \S 7.3 contains two theorems that confirm this.

The second of these two observations suggests that the expected difference between the makespan of a random schedule and the makespan of an optimal schedule is maximized at a value of $\frac {N} {M}$ somewhere in the neighborhood of 1.  This observation is particularly interesting in light of the empirical fact that square instances of the JSP (i.e., those with $\frac {N} {M} = 1$) are harder to solve than rectangular ones \cite {fisher63}.  

Figure 9 (B) graphs the number of search tree nodes ($90^{th}$ percentile) required by the branch and bound algorithm of Brucker et al. (1994) to optimally solve random $N$ by $M$ instances, as a function of the log (base 10) of search space size.  We take the size of the search space for an $N$ by $M$ JSP instance to be the number of possible disjunctive graphs, namely $2^{N {M \choose 2}}$.  Note that some of these disjunctive graphs contain cycles and therefore do not correspond to feasible schedules, so this expression overestimates the size of the search space.  Data points are given for each combination of $N$ and $M$ for which we could afford to run branch and bound (i.e., each combination of $N$ and $M$ for which we computed quantity (C)).  The data points are grouped into curves according to $\frac {N} {M}$.

Examining Figure 9 (B), we see that the curves are steepest for the ratios $\frac {2} {3}$, $1$, $\frac {3} {2}$, $2$, and $3$, and that the curves are substantially less steep for extreme values of $\frac {N} {M}$ such as $\frac {1} {7}$ and $7$.  Thus, at least from the point of view of this particular branch and bound algorithm, random JSP instances exhibit an ``easy-hard-easy" pattern of instance difficulty.  We discuss this pattern further in \S 7.4.

\subsection {Analysis}

The following two theorems show that, as $\frac {N} {M}$ approaches either $0$ or $\infty$, a random schedule will almost surely be near-optimal.

\begin {theorem} \label {randopt:0}
	Let $I$ be a random $N$ by $M$ JSP instance with optimal makespan $\ell_{min}(I)$ and let $S$ be a random schedule for $I$.  Then for fixed $N$ and $\epsilon > 0$, it holds whp (as $M \rightarrow \infty$) that $\ell(S) \le (1+\epsilon) \ell_{min}(I)$.
\end {theorem}
\begin {proof}
	The priority rule $\pi_{rand}$ associates a priority with each operation $o \in ops(I)$.  Let the sequence $T$ contain the elements of $ops(I)$, sorted in ascending order of priority.  The schedule $S = \mathcal {S} (\pi_{rand}, I)$ depends only on $T$, and there are $N M !$ possible choices of $T$.  Thus $\pi_{rand}$ can be seen as choosing at random from a set of $N M!$ instance-independent priority rules.  Because each of these instance-independent priority rules is subject to Lemma \ref {lem:backbone:0}, $\pi_{rand}$ is also subject to Lemma \ref {lem:backbone:0} and thus for each $J$, $\mathbb {E} [ \Delta^S_J ]$ is $O(N)$.  Thus $\mathbb {E} [\ell(S) - \ell_{min}(I)] \le \sum_J \mathbb {E} [\Delta^S_J] = O(N^2)$, so $\ell(S) - \ell_{min}(I)$ does not exceed $\epsilon \ell_{min}(I) = \Omega(M)$ whp by Markov's inequality.
\end {proof}

\begin {theorem} \label {randopt:inf}
	Let $I$ be a random $N$ by $M$ JSP instance with optimal makespan $\ell_{min}(I)$ and let $S$ be a random schedule for $I$.  Then for fixed $M$ and $\epsilon > 0$, it holds whp (as $N \rightarrow \infty$) that $\ell(S) \le (1+\epsilon) \ell_{min}(I)$.
\end {theorem}
\begin {proof}
See Appendix A.
\end {proof}

The idea behind the proof of Theorem \ref {randopt:inf} is the following.  As shown in Lemma \ref {lem:backbone:inf}, the priority rule $\pi_{\infty}$ almost surely generates an optimal schedule.  The relevant property of $\pi_{\infty}$ was that, when the operations were sorted in order of ascending priority, the number of operations in between $\mathcal {J}(o)$ and $o$ was $\Omega(N)$.  The key to the proof of Theorem \ref {randopt:inf} is that in expectation, $\pi_{rand}$ shares this property for \emph {most} of the operations $o \in ops(I)$.

\subsection {Easy-hard-easy Pattern of Instance Difficulty}

The proofs of Corollary \ref {optmakespan:0} (resp. Lemma \ref {lem:backbone:inf}) show that as $\frac {N} {M} \rightarrow 0$ (resp. $\frac {N} {M} \rightarrow \infty$) there exist simple priority rules that almost surely produce an optimal schedule.  Moreover, Theorems \ref {randopt:0} and \ref {randopt:inf} show that in these two limiting cases, even a random schedule will almost surely have makespan that is very close to optimal.  Thus, both as $\frac {N} {M} \rightarrow 0$ and as $\frac {N} {M} \rightarrow \infty$, almost all JSP instances are ``easy".

In contrast, for $\frac {N} {M} \approx 1$, Figure 9 (A) suggests that random schedules (as well as random local optima) are far from optimal.  The literature on the JSP (as well the results depicted in Figure 9 (B)) attests to the fact that random JSP instances with $\frac {N} {M} \approx 1$ are ``hard".  Thus we conjecture that, as in 3-SAT, typical instance difficulty in the JSP follows an ``easy-hard-easy" pattern as a function of a certain parameter.  In contrast to 3-SAT, the ``easy-hard-easy" pattern in the JSP is not (to our knowledge) associated with a phase transition (i.e., we have not identified a quantity that undergoes a sharp threshold at $\frac {N} {M} \approx 1$).

Furthermore, although the empirical results in Figures 9 (A) and (B) support the idea that typical-case instance difficulty in the JSP follows as ``easy-hard-easy" pattern, we do not claim to have isolated any particular value of $\frac {N} {M}$ as being the point of maximum difficulty.  As shown in Figure 9 (B), random JSP $N$ by $M$ JSP instances are most difficult for the branch and bound algorithm of Brucker et. al (1994) when $\frac {N} {M} \approx 2$, but this may not be true of other branch and bound algorithms or of JSP heuristics based on local search.  We leave the task of characterizing the ``easy-hard-easy" pattern more precisely as future work.

In related work, Beck (1997) \nocite {beck97} studied a constraint-satisfaction (as opposed to makespan-minimization) version of the JSP, and gave empirical evidence that the probability that a random JSP instance is satisfiable undergoes a sharp threshold as a function of a quantity called the \emph {constrainedness} of the instance.

\section {Limitations and Extensions}

The primary limitation of the work reported in this paper is that both our theoretical and empirical results apply only to \emph {random} instances of the job shop scheduling problem.  There is no guarantee that our observations will generalize to instances drawn from distributions with more interesting structure \cite {watson02}.  The difficulty in extending our analysis to other distributions is that analytical results similar to the ones presented in this paper may become much more difficult to derive.  However, there are at least three distributions that have been studied in the scheduling literature for which we believe it should be not too difficult to adapt our proofs (the conclusions may change as part of the adaptation process).
\begin {itemize}
	\item \emph {Random workflow JSP instances}.  In a workflow JSP instance, the set of machines is partitioned into sets (say $\mathcal{M}_1, \mathcal{M}_2, \ldots, \mathcal{M}_k$).  For $i<j$, each job must use all the machines in $\mathcal{M}_i$ before using any machines in $\mathcal{M}_j$.  Mattfeld et al. (1999) define a random distribution over workflow JSPs which generalizes in a natural way the distribution defined in \S 3.3 (the difference is that the permutations $\phi_1, \phi_2, \ldots, \phi_N$ are chosen uniformly at random from the set of permutations that satisfy the workflow constraints). \nocite {mattfeld99}
	\item \emph {Random instances of the (permutation) flow shop scheduling problem.}  An instance of the flow shop scheduling problem (FSP) is a JSP instance in which all jobs use the machines in the same order (equivalently, a FSP instance is a workflow JSP instance with $k=M$).  The \emph {permutation} flow shop problem (PFSP) is a special case of the FSP in which, additionally, each machine must process the jobs in the same order.  There is a large literature on the (P)FSP; Framinan et al. (2004) and Hejazi and Saghafian (2005) provide relevant surveys. \nocite {framinan04,hejazi05} 
	\item \emph {Job-correlated and machine-correlated JSP instances.}  In a \emph {job-correlated} JSP instance, the distribution from which operation durations are drawn depends on the job to which an operation belongs.  Similarly, in \emph {machine-correlated} JSP instance the distribution depends on the machine on which the operation is performed.  Watson et al. (2002) have studied job-correlated and machine-correlated instances of the PFSP.
\end {itemize}

Regarding the difficulty of instances drawn from these three distributions, computational experience shows that (i) random workflow JSPs are harder than random JSPs; (ii) random PFSPs are easier than random JSPs; and (iii) job-correlated and machine-correlated PFSPs are easier than random PFSPs.  Extending our theoretical analysis to each of these distributions may give some insight into the relevant differences between them.

\subsection {The Big Valley vs. Cost-Distance Correlations}

In \S 6, we defined a ``big valley" landscape as one that exhibits two properties: ``small improving moves" and ``clustering of global optima".  Our analytical and experimental results were based on this definition.  Although we believe this definition captures properties of JSP landscapes that are important for designers of heuristics to understand, other properties (e.g., cost-distance correlations) are likely to be important as well.  In particular, it may be possible for algorithms to exploit cost-distance correlations on landscapes that have neither the ``small improving moves" nor the ``clustering of global optima" properties.  

In the existing literature, the term ``big valley" is used amorphously to mean either  (1) a landscape like that depicted in Figure 1 or (2) a landscape that exhibits high cost-distance correlations.  By making a sharper distinction between these two distinct concepts, we can only improve our understanding of JSP landscapes as well as the landscapes of other combinatorial problems.

\section {Conclusions}

\subsection {Summary of Experimental Results}

Empirically, we demonstrated that for low values of the job to machine ratio ($\frac {N} {M}$), low-makespan schedules are clustered in a small region of the search space and the backbone size is high. As $\frac {N} {M}$ increases, low-makespan schedules become dispersed throughout the search space and the backbone vanishes.  As a function of $\frac {N} {M}$, the ``smoothness" of the landscape (as measured by a statistic called neighborhood exactness) starts out small for low values of $\frac {N} {M}$ (e.g., $\frac {N} {M} = \frac {1} {5}$), is relatively high for $\frac {N} {M} \approx 1$, and becomes small again for high values of $\frac {N} {M}$ (e.g., $\frac {N} {M} = 5$).  For both extremely low and extremely high values of $\frac {N} {M}$, the expected makespan of random schedules comes very close to that of optimal schedules.  The quality of random schedules (resp. random local optima) appears to be the worst at a value of $\frac {N} {M} \approx 1$.

\S 6.4 discussed the implications of our results for the ``big valley" picture of JSP search landscapes.  For $\frac {N} {M} \approx 1$, we concluded that a typical landscape can be described as a big valley, while for larger values of $\frac {N} {M}$ (e.g., $\frac {N} {M} \ge 3$) there are many big valleys.  \S 7.4 discussed how our data support the idea that JSP instance difficulty exhibits an ``easy-hard-easy" pattern as a function of $\frac {N} {M}$. 

\subsection {Summary of Theoretical Results}

Table 2 shows the asymptotic expected values of various attributes of a random $N$ by $M$ JSP instance in the limiting cases $\frac {N} {M} \rightarrow 0$ and $\frac {N} {M} \rightarrow \infty$.

\begin {center}
\begin {tabular} {c}
Table 2. Attributes of random JSP instances. \\
\begin {tabular} {| p{6.5cm} | p{3cm} | p{4cm} |}
	\hline 
	& Fixed $N$, $M \rightarrow \infty$	& Fixed $M$, $N \rightarrow \infty$	 \\[6pt]  \hline 
	Optimum makespan	& Max. job length (Corollary \ref {optmakespan:0})	& Max. machine workload (Corollary \ref{optmakespan:inf}) \\[11pt] \hline
	Normalized backbone size 	&	1 (Theorem \ref{backbone:0})	&	0 (Theorem \ref {backbone:inf}) \\[13pt]  \hline 
	Normalized maximum distance between global optima & 0 (Theorem \ref {backbone:0}) & $\Omega(1)$ (Theorem \ref{bigvalley:inf}) \\[11pt] \hline
	Normalized distance between random schedule and nearest global optimum	&	0 (Lemma \ref{closeness:0})	&	0 (Lemma \ref{closeness:inf}) \\[11pt] \hline
	Ratio of makespan of random schedule to optimum makespan	&	1 (Theorem \ref{randopt:0})	&	1 (Theorem \ref{randopt:inf}) \\[11pt] \hline
\end {tabular} \\
\end {tabular}
\end {center}

\subsection {Rules of Thumb for Designing JSP Heuristics}

Though we do not claim to have any deep insights into how to solve random instances of the JSP, our results suggest two general rules of thumb:
\begin {itemize}
	\item when $\frac {N} {M}$ is low (say, $\frac {N} {M} \approx 1$ or lower), an algorithm should attempt to locate the cluster of global optima and exploit it; while  
	\item when $\frac {N} {M}$ is high (say, $\frac {N} {M} \ge 3$) an algorithm should attempt to isolate one or more clusters of global optima and deal separately with each of them.
\end {itemize}
	
We briefly discuss these ideas in relation to two recent algorithms: \emph {backbone-guided local search} \cite {zhang04} and $i$-TSAB \cite {nowicki05}.

\subsubsection {Backbone-guided local search}

Several recent algorithms attempt to use backbone information to bias the move operator employed by local search.  For example, Zhang (2004) \nocite {zhang04} describes an approach called backbone-guided local search in which the frequency with which an attribute (e.g., an assignment of a particular value to a particular variable in a Boolean formula) appears in random \emph {local} optima is used as a proxy for the frequency with which the attribute appears in \emph {global} optima.  The approach improved the performance of the WalkSAT algorithm \cite {selman92} on large instances from SATLIB \cite {hoos00}.  A similar algorithm has been successfully applied to the TSP \cite {zhang05} to improve the performance of an iterated Lin-Kernighan algorithm \cite {martin91}.  Zhang writes:

\begin {quote}
This method is built upon the following working hypothesis: On a problem whose optimal and near optimal solutions form a cluster, if a local search algorithm can reach close vicinities of such solutions, the algorithm is effective in finding some information of the solution structures, backbone in particular. (Zhang, 2004, p. 3) \nocite {zhang04}
\end {quote}

Based on the results of \S\S4-5, this working hypothesis is satisfied for random JSPs with $\frac {N} {M} \approx 1$ or lower.  It seems plausible that backbone-guided local search could be used to boost the performance of early local search heuristics for the JSP such as those of van Laarhoven et al. \citeyear {vanLaar92} and Taillard \citeyear {taillard94} (whether the results would be competitive with those of recent algorithms such as $i$-TSAB is a separate question).  

The hypothesis is typically violated for random JSP instances with larger values of $\frac {N} {M}$.  In these cases it makes more sense to attempt to exploit \emph {local} clustering of optimal and near-optimal schedules.

\subsubsection {$i$-TSAB}

Nowicki and Smutnicki (2005) present a JSP heuristic called $i$-TSAB which employs multiple runs of the tabu search algorithm TSAB \cite {nowicki96}.  $i$-TSAB employs \emph {path relinking} to ``localize the center of BV [big valley], probably close to the global minimum" \cite {nowicki05}.  In other words, $i$-TSAB was designed based on the intuitive picture depicted in Figure 6 (A), which is inaccurate for typical random JSP instances with $\frac {N} {M} \ge 3$.   Note that although random JSP instances become ``easy" as $\frac {N} {M} \rightarrow \infty$, instances with $\frac {N} {M} \approx 3$ are by no means easy, as evidenced by Figure 9 (B).

For concreteness, we briefly describe how $i$-TSAB works.  Initially, $i$-TSAB performs a number of independent runs of TSAB and adds each best-of-run schedule to  a pool of ``elite solutions".  It then performs additional runs of TSAB and uses the best-of-run schedules from these additional runs to replace schedules in the pool of elite solutions.  Starting points for the additional TSAB runs are either (i) random elite solutions or (ii) schedules obtained by performing path relinking on a random pair of elite solutions.  Given two schedules $S_1$ and $S_2$, path relinking uses a move operator to generate a new schedule that is midway (in terms of disjunctive graph distance) between $S_1$ and $S_2$.  The pool of elite solutions can be thought of as a cloud of particles that hovers over the search space and (hopefully) converges to a region of the space containing a global optimum.

For random JSP instances with $\frac {N} {M} \approx 1$, our results are consistent with the idea that the cloud of elite solutions converges to the ``center" of the big valley.  For random JSP instances with $\frac {N} {M} \ge 3$, however, the cloud must either converge to one of many big valleys or not converge at all.  As an alternate approach one can imagine using multiple clouds, with the intention that each cloud specializes on a particular big valley.  It seems plausible that such ideas could improve the performance of $i$-TSAB on random JSP instances with larger values of $\frac {N} {M}$.

\section* {Appendix A: Additional Proofs}


For the proofs in this section, we define $\tau(O) \equiv \sum_{o \in O} \tau(o)$, where $O$ is any set of operations.  We make use of the following inequality \cite {spencer}.
\begin {api}
	Let $X = (X_1, X_2, \ldots, X_n)$ be a vector of $n$ independent random variables.  Let the function $f(x)$ take as input a vector $x = (x_1, x_2, \ldots, x_n)$, where $x_i$ is a realization of $X_i$ for $i \in [n]$, and produce as output a real number.  Suppose that for some $\beta > 0$ it holds that for any two vectors $x$ and $x'$ that differ on at most one component,
\begin {displaymath}
	| f(x) - f(x') | \le \beta \mbox { .}
\end {displaymath}

Then for any $\alpha > 0$, 
\begin {displaymath}
		\mathbb {P} \left [ X > \mathbb {E} [X] + \alpha \sqrt n \right ] \le \exp\left (- \frac {\alpha^2} {2 \beta^2} \right ) \mbox { .}
\end {displaymath}
\end {api}
The same inequality holds for $\mathbb {P} \left [ X \le \mathbb {E} [X] - \alpha \sqrt n \right ]$.


\begin {lemma2}
Let $I$ be a random $N$ by $M$ JSP instance.  Then for fixed $M$, it holds whp (as $N \rightarrow \infty$) that the schedule $S = \mathcal {S} (\pi_\infty, I)$ has the property that 
\begin {displaymath}
	S(o) = S^+(\mathcal {M} (o)) \mbox { } \forall o \in ops(I) \mbox { .} 
\end {displaymath}
\end {lemma2}

\begin {proof}

It remains only to prove Claim 2.1 from the proof in \S 4, which says that whp, for all $o \in ops^{2+}(I)$ we have 
\begin {displaymath}
S(o) - S(\mathcal {J} (o)) \ge \frac {1} {2} \mathbb {E} [S(o) - S(\mathcal {J} (o))] \mbox { .}
\end {displaymath}

Pick some arbitrary operation $o \in ops^{2+}(I)$, and suppose that the random choices used to construct $I$ were made in the following order:
\begin {enumerate}
	\item Randomly choose $m_1 = m(o)$ and $m_2 = \mathcal {J} (o)$.
	\item For $k$ from 1 to $N$:
	\begin {enumerate}
		\item Randomly choose the order in which job $J^k$ uses the machines (if $o \in J^k$ then part of this choice has already been made in step 1).
		\item Randomly choose $\tau(J^k_i)$ $\forall i \in [M]$.
	\end {enumerate}
\end {enumerate}

Let the random variable $X_k$ denote the sequence of random bits used in steps (a) and (b) of the $k^{th}$ iteration of the loop.  Define $\Delta_o \equiv S(o) - S(\mathcal {J} (o))$.  Then, for any fixed choices of $m_1$ and $m_2$, $\Delta_o$ is a function of the $N$ independent events $X_1, X_2, \ldots, X_N$, and it is easy to check that altering a particular $X_i$ changes the value of $\Delta_o$ by at most $2 \tau_{max}$.  Thus
\[
\begin {array} {l l}
\mathbb {P} \left [ \Delta_o < \frac {1} {2} \mathbb {E} [ \Delta_o ] \right ] & = \mathbb {P} \left [ \Delta_o < \mathbb {E} [ \Delta_o ] - \frac {\mu (N-1)} {2 M} \right ] \\
& \le \mathbb {P} \left [ \Delta_o < \mathbb {E} [ \Delta_o ] - \frac {\mu N} {2 M} \right ] \\
& \le \exp \left (- \frac {\mu^2 N} { 2 (4 M  \tau_{max})^2 }  \right )
\end {array}
\]
where in the first step we have used the fact (from the proof in \S 4) that $\mathbb {E} [\Delta_o] = \frac {\mu (N-1)} {M}$ and in the last step we have used A.P.I.  Taking a union bound over the $N (M-1)$ operations in $ops^{2+}(I)$ proves the claim.



\end {proof}


\begin {lemma3}
	Let $I$ be a random $N$ by $M$ JSP instance, and let $S$ be a random schedule for $I$.  Let $\hat S$ be an optimal schedule for $I$ such that $\| S - \hat S \|$ is minimal.  Let $f(M)$ be any unbounded, increasing function of $M$.  Then for fixed $N$, it holds whp (as $M \rightarrow$ $\infty$) that $\| S - \hat S \| < f(M)$.
\end {lemma3}

\begin {proof}
Let $\bar S  = \mathcal {S} (\pi_0, I)$.  The proof of Corollary \ref {optmakespan:0} showed that for any $J$, $\mathbb {E} [\Delta^{\bar S}_J]$ is $O(N^2)$.  Thus it holds whp that $\Delta^{\bar S}_J < \log(f(M))$ $\forall J$.  As in the proof of Theorem 5, the procedure used to produce $S$ is a mixture of instance-independent priority rules, each subject to Lemma \ref {lem:backbone:0}.  Thus for any $J$, $\mathbb {E} [\Delta^S_J$] is $O(N)$, so whp $\Delta^S_J < \log(f(M))$ $\forall J$.

Let $O_{near}(J_i) = \{ J'_j: J' \neq J, | \sum_{i'<i} \tau(J_{i'}) - \sum_{j'<j} \tau(J'_{j'}) | < \log(f(M)) \}$.  ($O_{near}(J_i)$ is the set of operations that would be scheduled ``near" in time to $J_i$ if we ignored the fact that a machine may only perform one operation at a time.)  Let $E_{near} = \{ e = \{ J_i, J'_j \} \in E(I): J'_j \in O_{near}(J_i)\}$.  Under the assumptions of the previous paragraph (each of which hold whp), $\| S - \bar S \| \le | E_{near} |$.  For any $J_i$, $\mathbb{E} [| O_{near}(J_i) | ]$ is $O(N \log f(M) )$.  Thus
\begin {displaymath}
	\mathbb {E} \left [\| S - \bar S \| \right ] \le \mathbb {E} \left [ |E_{near}| \right ] = \sum_{o \in ops(I)} \frac {1} {M} \mathbb {E} \left [ | O_{near} (o)| \right ] = O \left ( N^2 \log(f(M)) \right )
\end {displaymath}
so $\| S - \hat S \| < f(M)$ whp by Markov's inequality.
\end {proof}

For the purpose of the remaining proofs, it is convenient to introduce some additional notation.  Let $T = (T_1, T_2, \ldots, T_{|T|})$ be a sequence of operations.  We define
\begin {itemize}
	\item $T_{(i_1, i_2]} \equiv \{T_i: i_1 < i \le i_2\}$, and
	\item $T^{\bar m}_{(i_1, i_2]} \equiv \{ T_i \in T_{(i_1, i_2]}: m(T_i) = \bar m \}$ .
\end {itemize}


\begin {lemma4}
	Let $I$ be a random $N$ by $M$ JSP instance, let $S$ be a random schedule for $I$, and let $\hat S$ be an optimal schedule for $I$ such that $\| S - \hat S\|$ is minimal.  Then for fixed $M$ and $\epsilon > 0$, it holds whp (as $N \rightarrow$ $\infty$) that $\| S - \hat S\| < N^{1+\epsilon}$.
\end {lemma4}

\begin {proof}
	Let $T$ be the sequence of operations $o \in ops(I)$, sorted in ascending order by priority $\pi_{rand} (I,o)$ (where $\pi_{rand}$ is the random priority rule used to create $S$).  Note that for any $o \in ops(I)$ with $\mathcal {J} (o) \neq o^\emptyset$, $\mathcal {J} (o)$ must appear before $o$ in $T$.  Let $T_i$ denote the $i^{th}$ operation in $T$.  
	
	Consider the schedule $\bar S$ defined by the following procedure:

\begin {enumerate}
	\item $\bar S(o) \leftarrow \infty$ $\forall o \in ops(I)$.
	\item $Q \leftarrow ()$.  Let $Q_j$ denote the $j^{th}$ operation in $Q$.
	\item Let the function $ready(o)$ return true if $\bar S^+(\mathcal {M} (o)) \ge \bar S^+(\mathcal {J} (o))$, false otherwise.
	\item For $i$ from 1 to $N M$ do:
	\begin {enumerate}
		\item If $ready(T_i)$, then set $\bar S(o) \leftarrow \bar S^+(\mathcal {M} (T_i))$.  Otherwise append $T_i$ onto $Q$.
		\item For $j$ from 1 to $|Q|$ do:
		\begin {enumerate}
			\item If $ready(Q_j)$, then set $\bar S(Q_j) \leftarrow \bar S^+(\mathcal {M} (Q_j))$ and remove $Q_j$ from $Q$.
		\end {enumerate}
	\end {enumerate}
	\item Schedule any remaining operations of $Q$ in a manner to be specified (in the last paragraph of the proof).
\end {enumerate}
	
	The construction of $\bar S$ is just like the construction of $S$, except for the manipulations involving $Q$.  The purpose of $Q$ is to delay the scheduling of any operation $o$ that, if scheduled immediately, might produce a schedule in which $\bar S(o) > \bar S^+(\mathcal {M} (o))$. We first show that $\| S - \bar S\| < N^{1+\epsilon}$ whp; then we show that $\bar S$ is optimal whp.

	Let $Q^i$ denote $Q$ as it exists after $i$ iterations of step 4 have been performed.  Let $q(o) = \sum_{i=1}^{N M} |o \cap Q^i |$ be the number of iterations during which $o \in Q$.   We claim that $\| S - \bar S\| \le \sum_{o \in ops(I)} q(o) + (N-1) |Q^{N M}|$.  Letting $E^{\neq} = \{ e \in E(I): \vec e(\bar S) \neq \vec e(S) \}$, we have
	
\[
\begin {array} {l l}
	\| S - \bar S\| 	& = | \{ e \in E^{\neq}: e \cap Q^{N M} = \emptyset \} | + | \{ e \in E^{\neq}: e \cap Q^{N M} \neq \emptyset \} | \\
				& \le | \{ e \in E^{\neq}: e \cap Q^{N M} = \emptyset \} | + (N-1) | Q^{N M} | 
\end {array}
\]
	
	so it suffices to show $| \{ e \in E^{\neq}: e \cap Q^{N M} = \emptyset \} | \le \sum_{o \in ops(I)} q(o)$.  To see this, let $e = \{ o_1, o_2 \} \in E^{\neq}$ be such that $e \cap Q^{N M} = \emptyset$.  We must have $q(o_1) + q(o_2) > 0$.  We charge $e$ to the operation in $\{o_1, o_2\}$ that was inserted into $Q$ first.  It is easy to see that an operation can be charged for at most one edge per iteration it spends in $Q$, establishing our claim.  Thus it suffices to show that $\| S - \bar S\| \le \sum_{o \in ops(I)} q(o) + (N-1) |Q^{N M}| \le N^{1+\epsilon}$ whp.

	We divide the construction of $S$ into $n = M N^{\frac {1} {2} -  \epsilon'}$ epochs, each consisting of $N^{\frac {1} {2} + \epsilon'}$ iterations of step 4, for a to-be-specified $\epsilon' > 0$.  Let $z_j$ denote the number of iterations of step 4 that occur before the end of the $j^{th}$ epoch, with $z_{j} = 0$ for $j \le 0$ by convention.  Let
	
\begin {itemize}
	\item $C^{\bar m}_j \equiv T^{\bar m}_{(0, z_j]} \setminus Q^{z_j}$ be the set of operations that have been scheduled to run on $\bar m$ by the end of the $j^{th}$ epoch; and
	\item $O_{near} \equiv \bigcup_{j \in [n]} \{o \in T_{(z_{j-1}, z_j]}: \mathcal {J} (o) \in T_{(z_{j-(M+2)}, z_j]} \}$ be the set of operations whose job-predecessor belongs to a nearby epoch.
\end {itemize}
	
	For any $i \in [N M]$, $\mathbb {P} [ T_i \in O_{near} ] \le (M+2)N^{-\frac {1} {2} + \epsilon'}$.  Thus for any $j \in [n]$, $\mathbb {E} [ | O_{near} \cap T_{(z_{j-1}, z_j]} | ] \le (M+2)N^{2 \epsilon'}$.  Using A.P.I. it is straightforward to show that whp,
	
\begin {equation} \label {eq:Onear}
	| O_{near} \cap T_{(z_{j-1}, z_j]}| \le N^{\frac {1+\epsilon'} {2}} \mbox { } \forall j \in [n] \mbox { .}
\end {equation}
	
	We claim that whp, the following statements hold $\forall j \in [n]$:
	
\begin {align} 
	& \bigcup_{i \le z_j} Q^i \subseteq O_{near} \mbox { ,} \label {eq:8.2} \\
	& J \cap Q^{z_{j-1}} \neq \emptyset \Rightarrow |J \cap Q^{z_{j-1}} \cap Q^{z_j}| < |J \cap Q^{z_{j-1}}| & \forall J \in I \mbox { ,} \label {eq:8.3}\\
	& Q^{z_j} \cap Q^{z_{j-M}} = \emptyset \mbox { , and} \label {eq:8.4} \\
	& | Q^{z_j}| \le M N^{\frac {1+\epsilon'} {2}} \mbox { .} \label {eq:8.5}
\end {align}

	We prove this by induction, where each step of the induction fails with exponentially small probability.  For $j = 0$, \eqref {eq:8.3} and \eqref {eq:8.4} hold trivially.  \eqref {eq:8.2} is true because the operations in $T_{(0, z_1]} \setminus O_{near}$ are the first operations in their jobs, hence cannot be added to $Q$.  \eqref {eq:8.5} then follows from \eqref {eq:8.2} and \eqref {eq:Onear}.
	
	Consider the case $j > 0$.  To show \eqref {eq:8.2}, let $o$ be an arbitrary operation in $T_{(z_{j-1}, z_j]} \setminus O_{near}$.  By the induction hypothesis (specifically, equation \eqref {eq:8.4}), $\mathcal {J} (o) \in C^{m(\mathcal {J} (o))}_{j-2}$.  Thus $q(o) > 0 \Rightarrow \tau \left ( C^{m(\mathcal {J}(o))}_{j-2} \right ) > \tau \left ( C^{m(o)}_{j-1} \right )$.  By the induction hypothesis,
	
\begin {displaymath}
	\tau \left ( C^{m(o)}_{j-1} \right ) - \tau \left ( C^{m(\mathcal {J}(o))}_{j-2} \right ) \ge \tau \left ( T^{m(o)}_{(0, z_{j-1}]} \right ) - M N^{\frac {1+\epsilon'} {2}} - \tau \left ( T^{m(\mathcal {J} (o))}_{(0, z_{j-2}]} \right ) \mbox { .}
\end {displaymath}
	
	Letting $\Delta$ denote the right hand side of this inequality, we have $\mathbb {E} [\Delta] = \frac {1} {M} N^{\frac {1} {2} + \epsilon'} - M N^{\frac {1+\epsilon'} {2}}$, and A.P.I. can be used to show that for some $K > 0$ independent of $N$, $\mathbb {P} [\Delta < 0] \le \exp(- \frac {1} {K} N^{\epsilon'})$.  Thus \eqref {eq:8.2} holds with probability at least $1-\exp(- \frac {1} {K} N^{\epsilon'})$.
	
	To show \eqref {eq:8.3}, let $J$ be such that $J \cap Q^{z_{j-1}} \neq \emptyset$, and let $J_i \in Q^{z_{j-1}}$ be chosen so that $i$ is minimal.  Then $\mathcal {J} (J_i) \in C^{m(\mathcal {J} (J_i))}_{j-1}$.  Thus $J_i \in Q_{z_j} \Rightarrow \tau \left ( C^{m(\mathcal {J} (J_i))}_{j-1} \right ) > \tau \left ( C^{m(J_i)}_{j} \right )$.  By \eqref {eq:Onear}, \eqref {eq:8.2}, and the induction hypothesis (equation \eqref {eq:8.5}), $|Q^{z_j}| \le (M+1)N^{\frac {1+\epsilon'} {2}}$.  Using the same technique as above, we can show that \eqref {eq:8.3} holds with probability at least  $1-\exp(- \frac {1} {K} N^{\epsilon'})$ for some $K > 0$ independent of $N$.  
	
	\eqref {eq:8.3} implies \eqref {eq:8.4}.  \eqref {eq:8.2} and \eqref {eq:8.4} together with \eqref {eq:Onear} imply \eqref {eq:8.5}.  Thus whp, \eqref {eq:8.2} through \eqref {eq:8.5} hold $\forall j \in [n]$.  
	
	By \eqref {eq:8.2} and \eqref {eq:8.4}, we have
	
\begin {displaymath}
	\mathbb {E} \left [ \sum_{o \in ops(I)} q(o) \right ] \le \mathbb {E} [| O_{near} | ]  M N^{\frac {1} {2} + \epsilon'} \le M^2 (M+2) N^{1 + 2 \epsilon'}
\end {displaymath}

	and also
	
\begin {displaymath}
	\mathbb {E} [ | Q^{N M} |] \le \mathbb {E} [ | T_{(z_{n-M}, z_n]} \cap O_{near} | ] \le (M+2) N^{2 \epsilon'}
\end {displaymath}

	so setting $\epsilon' = \frac {\epsilon} {3}$ gives $\| S - \bar S\| \le \sum_{o \in ops(I)} q(o) + (N-1) | Q^{N M} | \le N^{1 + \epsilon}$ whp.

	It remains to show that $\bar S$ is optimal whp.  We first prove the following claim.
\\
\\
{\bf Claim 4.1.} For any non-negative integers $a$ and $b$, the probability that $T_{(a,b]}$ contains two operations from the same job is at most $\frac {(b-a)^2} {N}$. 
\begin {proof} [Proof of Claim 4.1]
Let $X$ denote the number of \emph {pairs} of operations in $T_{(a,b]}$ that belong to the same job.  Then $\mathbb {P} [X > 0] \le \mathbb {E} [X] \le {b-a \choose 2} \frac {1} {N} \le \frac {(b-a)^2} {N}$.  
\end {proof}

	To see that $\bar S$ is optimal whp, note that the operations scheduled prior to step 5 do not cause any idle time on any machine, so it is only the operations in $Q^{N M}$ that can cause $\bar S$ to be sub-optimal.  Let $\tau(\bar m)$ $\equiv$ $\tau(\{ o \in ops(I): m(o) = \bar m \})$ denote the workload of machine $\bar m$.  Let $\hat m = \argmax_{\bar m \in [M]} \tau(\bar m)$.  Then the following hold whp.

\begin {itemize}
	\item The set $Z^{\hat m} \equiv T^{\hat m}_{( NM-2 M N^{\frac {1} {4}}, NM ]}$ consists of operations belonging to jobs that use $\hat m$ last.  (It holds whp that $Z^{\hat m} \subset Z$, where $Z \equiv T_{(NM-N^{\frac {1} {3}},NM]}$.  So if $Z^{\hat m}$ contains an operation from a job that does not use $\hat m$ last, then $Z$ must contain two operations from the same job.  But by Claim 4.1, the probability that this happens is at most $(N^{\frac {1} {3}})^2 \frac {1} {N} = o(1)$.)
	\item $\mu N^{\frac {1} {4}} \le \tau (Z^{\hat m})$ and $\tau(Z^{\hat m}) \le \tau(\hat m) - \tau(\bar m)$ $\forall \bar m \neq \hat m$.  (This follows by applying the Central Limit Theorem to $\tau(Z^{\hat m})$, $\tau(\hat m)$, and $\tau(\bar m)$).
\end {itemize}

	Thus whp it holds that prior to the execution of step 5, $\bar S$ contains a period of length at least $\tau(Z^{\hat m}) \ge \mu N^{\frac {1} {4}}$ during which the only operations being processed are those in $Z^{\hat m}$, where $\{o \in ops(I): \mathcal {J} (o) \in Z^{\hat m} \} = \emptyset$.  Assuming $|Q^{N M}| < N^{3 \epsilon'}$ (holds whp), we can always schedule the operations in $Q^{N M}$ so as to guarantee $\ell(\bar S) = \tau(\hat m)$, which implies $\bar S$ is optimal.
	
\end {proof}


\begin {theorem6}
	Let $I$ be a random $N$ by $M$ JSP instance with optimal makespan $\ell_{min}(I)$ and let $S$ be a random schedule for $I$.  Then for fixed $M$ and $\epsilon > 0$, it holds whp (as $N \rightarrow \infty$) that $\ell(S) \le (1+\epsilon) \ell_{min}(I)$.
\end {theorem6}

\begin {proof}
	
As in the proof of Lemma 4, let $T$ be the sequence of operations $o \in ops(I)$, sorted in ascending order by priority $\pi_{rand} (I,o)$ (where $\pi_{rand}$ is the random priority rule used to create $S$).  Note that for any $o \in ops(I)$ with $\mathcal {J} (o) \neq o^\emptyset$, $\mathcal {J} (o)$ must appear before $o$ in $T$.  Let $T_i$ denote the $i^{th}$ operation in $T$.
	
Rather than analyze $S$ directly, we analyze a schedule $\bar S$ defined by the following procedure:
	
	\begin {enumerate}
		\item $t \leftarrow 0$.
		\item For $i$ from 1 to $N M$ do:
			\begin {enumerate}
				\item Set $\bar S(T_i) = \max (t, \bar S^+(\mathcal {J} (T_i)), \bar S^+(\mathcal {M} (T_i)))$ . 
				\item If $\bar S^+(\mathcal {J} (T_i)) > \bar S^+(\mathcal {M} (T_i))$, set $t = \max_{i' \le i} \bar S^+(T_{i'})$.
			\end {enumerate}
	\end {enumerate}
	
	The procedure is identical to the one used to construct $S$, except that, whenever an operation $T_i$ is assigned a start time $\bar S(T_i) > \bar S^+(\mathcal {M} (T_i))$, the procedure inserts artificial delays into the schedule in order to re-synchronize the machines.  For any $T$, it is clear that $\ell(S) \le \ell(\bar S)$.  Thus, it suffices to show that $\ell(\bar S) \le (1+\epsilon) \ell_{min}(I)$ whp.
	
	We divide the construction of $\bar S$ into $n$ epochs, where each update to $t$ (in step 2(b)) defines the beginning of a new epoch.  Let $z_i$ be the number of operations scheduled before the end of the $i^{th}$ epoch, with $z_0 = 0$ by convention.  Let $t_i = \max_{i' \le z_i} S^+(o_{i'})$ be the (updated) value of $t$ at the end of the $i^{th}$ epoch.  Define $\Delta_i \equiv \sum_{\bar m=1}^M t_i - \max_{i' < i, m(T_{i'}) = \bar m} S^+(T_{i'})$.  Then $\ell (\bar S) - \ell_{min} (I) \le \sum_{i=1}^n \Delta_i$, so it suffices to show that $\sum_{i=1}^n \Delta_i \le \epsilon \ell_{min}(I)$ whp.
	
	Let $I = [n]$, and let $L = \{i \in I: z_i - z_{i-1} \ge N^{\frac {2} {7}} \}$.  We first consider $\sum_{i \in L} \Delta_i$; then we consider $\sum_{i \in I \setminus L} \Delta_i$.

	Let $i_1$ and $i_2$ be arbitrary integers with $0 \le i_1,i_2 \le N M$ and $i_2 - i_1 \ge N^{\frac {2} {7}}$.  Let $\bar \tau = \tau(T^{\bar m}_{(i_1, i_2]})$.  Then $\mathbb {E} [\bar \tau] = \mu \frac {i_2 - i_1} {M}$.  For any $T$, $\bar \tau$ is a function of the outcome of at most $i_2 - i_1$ events (namely, the definition of each of the jobs in $\{J: J \cap T_{(i_1, i_2]}\neq \emptyset \}$), each of which alters the value of $\bar \tau$ by at most $\tau_{max}$.  It follows by A.P.I. that
	
\begin {displaymath}
	\mathbb {P} [ | \bar \tau - \mathbb {E} [ \bar \tau ] | > N^{\epsilon'} \sqrt {i_2 - i_1} ] \le 2 \exp \left ( - \frac { N^{2 \epsilon'} } {2 \tau_{max}^2} \right )
\end {displaymath}
	
	for any $\epsilon' > 0$.  Thus, it holds whp that $| \bar \tau - \mathbb {E} [ \bar \tau ] | \le N^{\epsilon} \sqrt {i_2 - i_1}$ for all possible choices of $i_1$ and $i_2$.  In particular, whp we have $\Delta_i \le 2 M N^{\epsilon'} \sqrt {z_i - z_{i-1}} \le $ $\forall i \in L$, which implies $\sum_{i \in L} \Delta_i \le N^{\frac {5} {7}}  \sum_{i \in L} 2 M N^{\epsilon'} \sqrt {N^{\frac {2} {7}}} =   2 M N^{\frac {6} {7} + \epsilon'}$.
	
	Now consider $\sum_{i \in I \setminus L} \Delta_i$.  As shown in the proof of Lemma 4 (Claim 4.1), for any non-negative integers $a$ and $b$ the probability that $T_{(a,b]}$ contains two operations from the same job is at most $\frac {(b-a)^2} {N}$.  Thus the probability that an arbitrary subsequence of size $N^{\frac {2} {7}}$ contains two operations from the same job is at most $N^{- \frac {3} {7}}$, so $\mathbb {E} [ | I \setminus L |] \le N^{\frac {4} {7}}$.  Clearly $\Delta_i \le \tau_{max} N^{\frac {2} {7}}$ $\forall i \in I \setminus L$, so $\mathbb {E} [ \sum_{i \in I \setminus L} \Delta_i ]$ is $O(N^{\frac {6} {7}})$.

	Thus $\mathbb {E} [\sum_{i \in I} \Delta_i]$ is $O(N^{\frac {6} {7} + \epsilon'})$ for any $\epsilon' > 0$, so $\sum_{i \in I} \Delta_i \le N^{\frac {6} {7} + 2 \epsilon'}$ whp, while it is easy to see that  $\ell_{min}(I) \ge \mu \frac {N} {2}$ whp.
	 
\end {proof}

\vskip 0.2in
\bibliographystyle{theapa}
\bibliography{jair}

\end{document}